\documentclass[nonblindrev]{informs3} 

\usepackage{balance}
\usepackage{amsmath}
\usepackage[tight,footnotesize]{subfigure}
\usepackage{graphicx}
\usepackage{bm}
\usepackage{algorithm}
\usepackage[noend]{algorithmic}
\usepackage{multirow}
\usepackage{slashbox}
\usepackage{caption}
\usepackage{amsfonts}
\usepackage{siunitx}

\OneAndAHalfSpacedXII 

\usepackage{natbib}
 \bibpunct[, ]{(}{)}{,}{a}{}{,}%
 \def\bibfont{\small}%
\TheoremsNumberedThrough     

\EquationsNumberedThrough    

\MANUSCRIPTNO{2015} 

\begin{document}


\RUNAUTHOR{Tang and Yuan}

\RUNTITLE{Non-monotone Sequential Submodular Maximization}

\TITLE{Non-monotone Sequential Submodular Maximization}

\ARTICLEAUTHORS{%
\AUTHOR{Shaojie Tang}
\AFF{Naveen Jindal School of Management, The University of Texas at Dallas}
\AUTHOR{Jing Yuan}
\AFF{Department of Computer Science and Engineering, The University of North Texas}
} 

\ABSTRACT{
In this paper, we study a fundamental problem in submodular optimization, which is called sequential submodular maximization. Specifically, we aim to select and rank a group of $k$ items from a ground set $V$ such that the weighted summation of $k$ (possibly non-monotone) submodular functions $f_1, \cdots ,f_k: 2^V \rightarrow \mathbb{R}^+$  is maximized, here each function $f_j$ takes the first $j$ items from this sequence as input.
The existing research on sequential submodular maximization has predominantly concentrated on the monotone setting, assuming that the submodular functions are non-decreasing. However, in various real-world scenarios, like diversity-aware recommendation systems, adding items to an existing set might negatively impact the overall utility. In response, this paper pioneers the examination of the aforementioned problem with non-monotone submodular functions and offers effective solutions for both flexible and fixed length constraints, as well as a special case with identical utility functions. The empirical evaluations further validate the effectiveness of our proposed algorithms in the domain of video recommendations. The results of this research have implications in various fields, including  recommendation systems and assortment optimization, where the ordering of items significantly impacts the overall value obtained.}


\maketitle
\section{Introduction}
Submodular optimization is a central problem in machine learning with various applications in a wide range of fields, including data summarization \citep{lin2011class}, sparse reconstruction \citep{das2011submodular}, active learning \citep{golovin2011adaptive,tang2021optimal}, and viral marketing \citep{tang2020influence}. These formulations aim to select a subset of items that maximizes a submodular function. However, in many real-world applications, the objective is not only to select items but also to rank them in a specific order \citep{azar2011ranking,tschiatschek2017selecting,tang2021cascade}. This motivates the study of \emph{sequential submodular maximization} \citep{asadpour2022sequential,zhang2022ranking}, a fundamental problem in submodular optimization. This problem involves selecting and ranking a group of $k$ items from a ground set $V$. The goal is to maximize the weighted summation of $k$ submodular functions, denoted as $f_1, \cdots ,f_k: 2^V \rightarrow \mathbb{R}^+$, where each function $f_j$ takes the first $j$ items from the ranking sequence as input. Formally, the objective is to find a feasible sequence $\pi=\{\pi_1, \cdots, \pi_k\}$ over items in $V$ that maximizes the value of
\begin{eqnarray}
\label{eq:0}
F(\pi) \stackrel{\text{def}}{=} \sum_{j\in[k]}\lambda_j\cdot f_j(\pi_{[j]})
 \end{eqnarray} where $\lambda_j$ denotes the weight of function $j$ and $\pi_{[j]} \stackrel{\text{def}}{=} \{\pi_1, \cdots, \pi_j\}$ denotes the first $j$ items of $\pi$.

This problem, which captures the position-bias in item selection, has a wide range of applications, including sequential active learning and recommendation systems.  For instance, it can be applied to tackle challenges in ranking products on online retail platforms \citep{asadpour2022sequential}. Platforms like Amazon and Airbnb face the task of not only selecting a subset of products or rooms to showcase but also arranging them in a vertical list format to provide customers with an organized and customer-friendly browsing experience. Shoppers scroll through this list, depending on their level of patience, and may potentially make a purchase from the products displayed. The platform's objective is to optimize the selection and ranking of products to maximize the likelihood of a purchase. Interestingly, these applications can be framed as a problem of sequential submodular maximization. In this context, the parameters in $F(\pi)$  can be interpreted as follows: We denote the set of products as $V$, the window size of displayed prodcuts as $k$, $\lambda_j$ represents the proportion of customers with a specific patience level $j$ (e.g., a customer with patience level $j$ is willing to view the first $j$ products $\pi_{[j]}$). The function \(f_j(\pi_{[j]})\) denotes the likelihood of purchase for customers with a patience level of $j$ after seeing the first $j$ products $\pi_{[j]}$. Typically, $f_j$ is described as a submodular function. In this case,  $F(\pi)$  captures the expected probability of purchase when a customer is shown a sequence of products $\pi$.

 While previous research on sequential submodular maximization has primarily focused on the monotone setting, where submodular functions are assumed to be non-decreasing, the non-monotonicity of submodular functions becomes more apparent in many real-world scenarios, including feature selection \citep{das2008algorithms},  maximum cut \citep{gotovos2015non}, profit maximization \citep{tang2021adaptive}, and data summarization \citep{mirzasoleiman2016fast}.  One such example involves designing a \emph{diversity-aware} recommendation system for a vast assortment of products spanning different categories \citep{lin2010multi,mirzasoleiman2016fast,amanatidis2020fast,carbonell1998use}. The system's primary goal is to generate a sequence of products that not only have high ratings but also effectively represent the entire collection. To address this tradeoff, a commonly adopted objective function is submodular but not monotone. More details about this application can be found in the experiment section.  This highlights the need to develop algorithms that can handle non-monotone submodular functions efficiently.

We present the main contributions of this paper as follows:

1. This paper marks the first exploration of the sequential submodular maximization problem with non-monotone submodular functions. We investigate this problem with two constraints: a flexible length constraint, where we seek to compute a sequence of items with a length of at most $k$, and a fixed length constraint, where our objective is to compute a sequence of items with an exact length of $k$.

2. For the flexible length constraint, we introduce efficient constant-factor algorithms. Additionally, for the fixed length constraint, we develop an algorithm with an approximation ratio that depends on the ratio of $k/n$, where $n$ represents the size of the ground set.

3. In the scenario where all $k$ utility functions are identical, meaning there exists some function $f$ such that $f_i=f$ for all $j\in \{1, 2, \cdots, k\}$, we develop constant-factor approximation algorithms subject to the fixed length constraint.

4.  To evaluate the effectiveness of our algorithms in the context of video recommendations, we conduct comprehensive experiments using the Movielens dataset. The experimental results demonstrate the superiority of our proposed solutions, validating their efficacy and practicality.

\paragraph{Additional related work.} The traditional non-monotone submodular optimization, where the objective is to select a \emph{set} of items to maximize a non-monotone submodular function, has been extensively studied in the literature \citep{gharan2011submodular, buchbinder2014submodular}. The best-known result for this problem, subject to a cardinality constraint, is a $0.385$-approximation algorithm \citep{buchbinder2019constrained}. It is crucial to emphasize that our work differs from this traditional setting. Instead of aiming to find a \emph{set} of items where the order does not matter, our goal is to find a \emph{sequence} of items to maximize a group of submodular functions. It will become evident later that their problem can be seen as a special case of our setting. The focus on sequences introduces a new dimension of complexity in comparison to the traditional set-based approaches. However, we draw inspiration from previous studies \citep{amanatidis2020fast, tang2021beyond} and incorporate a sampling technique to tackle the challenges arising from non-monotonicity. While there are other studies that have explored position bias in submodular optimization \citep{tschiatschek2017selecting,alaei2010maximizing}, our problem formulation significantly differs from theirs.
\section{Preliminaries and Problem Statement}
\textbf{Notations.} We first introduce some useful notations. Throughout the remainder of this paper, we denote the set $\{1, 2, \ldots, m\}$ as $[m]$ for any positive integer $m$. Given a function  $f$, we use $f(i\mid S)$ to denote the marginal utility of adding $i$ to $S$, i.e., $f(i\mid S)  \stackrel{\text{def}}{=} f(S\cup\{i\})-f(S)$.
We say a function $f$ is submodular if and only if $f(i\mid X) \geq f(i\mid Y)$ for any two sets $X$ and $Y$ such that $X\subseteq Y$ and any item $i\notin Y$.
Let $\pi=\{\pi_1, \cdots, \pi_k\}$ be a sequence of items, we define the operation $\pi\oplus i$ as the concatenation of $i$ to $\pi$, that is, $\pi\oplus i \stackrel{\text{def}}{=} \{\pi_1, \cdots, \pi_k, i\}$.

Now we are ready to introduce our research problem. Given $k$ non-monotone submodular functions $f_1, \cdots ,f_k: 2^V \rightarrow \mathbb{R}^+$ and non-negative coefficients $\lambda_1, \cdots ,\lambda_k$, the objective of the Non-Monotone Sequential Submodular Maximization (NSM) problem is to find a \emph{feasible} sequence $\pi=\{\pi_1, \cdots, \pi_k\}$ over items in $V$ that maximizes the value of  $F(\pi)$. Here
\begin{eqnarray}
F(\pi) \stackrel{\text{def}}{=} \sum_{j\in[k]}\lambda_j\cdot f_j(\pi_{[j]})
 \end{eqnarray} where $\pi_{[j]} \stackrel{\text{def}}{=} \{\pi_1, \cdots, \pi_j\}$ denotes the first $j$ items of $\pi$. In this paper, we adopt a non-standard notation, employing $\pi$ to represent both a sequence of items and the set of items comprising this sequence. To simplify notation, define $\pi_j=\emptyset$ if $|\pi|<j$ where $|\pi|$ denotes the number of \emph{non-empty} items contained in $\pi$.

The set of feasible sequences is denoted by $\Pi$. Hence, our goal is to maximize the objective function $F(\pi)$ over all possible sequences $\pi$ in $\Pi$.  In this paper, we consider two types of feasibility constraints. The first type, known as NSM with Flexible Length, imposes a constraint where feasible sequences can contain at most $k$ items. In other words, $\Pi\stackrel{\text{def}}{=}  \{\pi \mid |\pi|\leq k\}$.
 The second type, known as NSM with Fixed Length, requires that all feasible sequences contain a fixed number $k$ of items from $V$. Formally, $\Pi\stackrel{\text{def}}{=}  \{\pi \mid |\pi|=k\}$ in this case.
Both problems have additional restrictions: the same item cannot appear multiple times in a feasible sequence, and there should be no empty slots between two items. These constraints ensure that each item is considered at most once and maintain the sequential nature of the sequence.


\section{NSM with Flexible Length}
\label{sec:flexible}
A formal description of NSM with Flexible Length can be written as follows:
 \begin{center}
\framebox[0.4\textwidth][c]{
\enspace
\begin{minipage}[t]{0.4\textwidth}
\small
$\textbf{P.1}$
$\max_{\pi}  F(\pi)$ subject to $|\pi|\leq k$.
\end{minipage}
}
\end{center}
\vspace{0.1in}

We first provide a negative result by showing that it is impossible to find a $0.491$-approximation algorithm for $\textbf{P.1}$. This result can be easily shown by setting $\lambda_1=\lambda_2=\cdots, =\lambda_{k-1}=0$ and $\lambda_k=1$ in $\textbf{P.1}$, thereby reducing the problem to maximizing a non-monotone submodular function $f_k$ over a \emph{set} of at most $k$ items. It is a well-known fact that maximizing a non-monotone submodular function subject to a cardinality constraint does not allow for a $0.491$-approximation solution \citep{gharan2011submodular}. Thus, we establish the following lemma.
\begin{lemma}
It is impossible to achieve a $0.491$-approximation for $\textbf{P.1}$.
\end{lemma}
\subsection{Algorithm design}

Next, we present the design of our algorithm, referred to as \textsf{Sampling-Greedy}, which builds upon a simple greedy approach that selects items based on their marginal utility. However, since our utility function is non-monotone, employing a straightforward greedy strategy may result in suboptimal selections with low utility. To address this challenge, we draw inspiration from \citep{amanatidis2020fast} and introduce a sampling phase to the greedy algorithm, extending its guarantees to the non-monotone setting. We provide a detailed description of our algorithm below, which consists of two phases:
\begin{enumerate}
 \item \textsc{Random Subset Selection:} We begin by selecting a random subset, denoted as $R$, from the ground set $V$. Each item $i\in V$ is independently included in $R$ with a probability $p\in[0,1]$, where $p$ is a parameter to be optimized later. This random subset serves as the initial pool of items for subsequent processing.

 \item \textsc{Greedy Algorithm on $R$:} We run a greedy algorithm only on $R$. This algorithm operates iteratively, augmenting the current sequence by selecting an item that provides the greatest incremental utility. To be precise, consider a specific round of the greedy algorithm, let $\pi$,  whose initial value is $\emptyset$, denote the current solution. Let $z\in \argmax_{i\in R} \left[ F(\pi\oplus i) - F(\pi) \right]$ denote the item that has the largest marginal utility on top of $\pi$. If $F(\pi\oplus z) - F(\pi) > 0$, then we append $z$ to $\pi$ (i.e., $\pi= \pi\oplus z$) and proceed to the next iteration. Note that $F(\pi\oplus j) - F(\pi)= \sum_{j\in\{|\pi|+1, \cdots, k\}} \lambda_j\cdot f_j(i\mid \pi)$, observing that appending $j$ to $\pi$ only affects the utility of those functions in $\{f_{|\pi|+1}, \cdots, f_k\}$.  This construction process continues until one of two stopping criteria is met: either the sequence reaches a length of $\min\{k, |R|\}$, or the marginal utility of the remaining items becomes non-positive. This ensures that further additions would not contribute positively to the overall utility.
\end{enumerate}


\begin{algorithm}[hptb]
\caption{\textsf{Sampling-Greedy}}
\label{alg:1}
\begin{algorithmic}[1]
\STATE $E=V$, $\pi=\emptyset$, $t=1$, $Q=\{i\in E\mid \sum_{j\in\{1, \cdots, k\}} \lambda_j\cdot f_j(\{i\}) > 0\}$
\WHILE {$t\leq k$ and $Q\neq \emptyset$}
\STATE \underline{consider} $z=\argmax_{i\in Q} \sum_{j\in\{t, \cdots, k\}} \lambda_j\cdot f_j(i\mid \pi)$  \label{line:3}
\STATE $E=E\setminus \{z\}$
\STATE let $\Phi_z \sim \mathrm{Bernoulli}(p)$
\IF {$\Phi_z=1$}
\STATE $\pi = \pi\oplus z$, $t\leftarrow t+1$
\STATE $Q=\{i\in E\mid \sum_{j\in\{t, \cdots, k\}} \lambda_j\cdot f_j(z\mid \pi) > 0\}$
\ENDIF
\ENDWHILE
\RETURN $\pi$
\end{algorithmic}
\end{algorithm}

In order to facilitate analysis, we present an alternative approach to implementing \textsf{Sampling-Greedy} (a detailed description of this alternative is listed in Algorithm \ref{alg:1}). Unlike the original implementation, where the entire set $R$ is sampled at the start of the algorithm, our alternative defers this decision. Instead, we employ a coin toss, with a success probability of $p$, after \emph{considering} each item. This coin toss determines whether the item should be added to the solution. It is straightforward to confirm that both versions of the algorithm produce same output distributions.

\subsection{Performance analysis}
We next analyze the approximation ratio of \textsf{Sampling-Greedy}.  As in Algorithm \ref{alg:1}, we introduce a random variable $\Phi\in \{0,1\}^V$ to denote the outcome  of the coin tosses, e.g., $\Phi_i\in\{0, 1\}$ represents the coin toss of item $i$.

Let $\pi^*=\{\pi^*_1, \cdots, \pi^*_l\}$ denote the optimal solution of $\textbf{P.1}$, where $l\leq k$.
In the context of a specific run of our algorithm, where the coin tosses $\Phi$ is realized and the corresponding sequence returned by our algorithm is denoted as $\pi(\Phi)$, we \emph{partition} the optimal solution $\pi^*$ into three sets as follows:

1. Set $O_{\textsf{cons.}}(\Phi)$: This set contains all items $\pi^*_j$ in the optimal sequence $\pi^*$ that have been considered by our algorithm before (including when) position $j$ being filled,  but were not picked due to the random coin flip $\Phi$. That is, $O_{\textsf{cons.}}(\Phi)\stackrel{\text{def}}{=}  \{\pi^*_j\in \pi^*  \mid \pi^*_j \mbox{ was considered by Algorithm \ref{alg:1}  during the selection of } \pi(\Phi)_{[j]} \mbox{ but } \pi^*_j \notin \pi(\Phi)_{[j]}\}$, where $ \pi(\Phi)_{[j]}$ denotes the first $j$ items of $\pi(\Phi)$.

2. Set $O_{\textsf{not cons.}}(\Phi)$: This set consists of all items $\pi^*_j$ in $\pi^*$ that have not been considered  during our algorithm before (including when) position $j$  being filled. That is, $O_{\textsf{not cons.}}(\Phi) \stackrel{\text{def}}{=}  \{\pi^*_j\in \pi^*  \mid \pi^*_j \mbox{ was not considered by Algorithm \ref{alg:1}  during the selection of } \pi(\Phi)_{[j]} \}$.

3. Set $O_{\textsf{ovlpd.}}(\Phi)=\pi^*\setminus (O_{\textsf{not cons.}}(\Phi)\cup O_{\textsf{cons.}}(\Phi))$: This set contains those items $\pi^*_j$ in $\pi^*$ that have been added to  $\pi(\Phi)$ before (including when) position $j$ being filled. That is, $O_{\textsf{ovlpd.}}(\Phi) \stackrel{\text{def}}{=}  \{\pi^*_j\in \pi^*  \mid \pi^*_j \in \pi(\Phi)_{[j]}\}$.

Let's consider a specific example to illustrate the aforementioned three sets. Suppose we have an optimal sequence $\pi^* = \{a, b, c, d\}$. And let's assume that the sequence considered by our algorithm is  $\{b, c, d, e, f, g\}$, and the final sequence picked by our algorithm, based on the coin tosses $\Phi$, is $\pi(\Phi) = \{c, e, f, g\}$. For this example, $O_{\textsf{cons.}}(\Phi)=\{b, d\}$, this is because our algorithm considered  $b$ (resp. $d$) before filling position  $2$ (resp. $4$), but did not pick $b$ (resp. $d$); $O_{\textsf{not cons.}}(\Phi)=\{a\}$, this is because our algorithm did not consider $a$ before filling position  $1$; $O_{\textsf{ovlpd.}}(\Phi)=\{c\}$, this is because our algorithm picked $c$  before filling position  $3$.

By defining this partition of $\pi^*$, we can effectively analyze the influence of the coin tosses $\Phi$ (or, equivalently, the random subset $R$ in the original implementation of our algorithm) on the resultant sequence and its connection to the optimal solution.

To simplify notation, we drop the random variable $\Phi$ from $\pi(\Phi)$, $O_{\textsf{cons.}}(\Phi)$, $O_{\textsf{not cons.}}(\Phi)$ and $O_{\textsf{ovlpd.}}(\Phi)$ if it is clear from the context.

Before analyzing the performance of our algorithm, we introduce some additional notations. Recall that we define $\pi_j=\emptyset$ if $j>|\pi|$. Given a random output $\pi$ from our algorithm and an optimal solution $\pi^*$, we define $F(\pi \uplus \pi^*)=\sum_{j\in[k]}\lambda_j\cdot f_j(\pi_{[j]}\cup\pi^*_{[j]})$ as the utility of the ``union'' of $\pi$ and $\pi^*$. Here, $\pi_{[j]}$ (resp. $\pi^*_{[j]}$) denotes all items from $\pi$ (resp. $\pi^*$) that are placed up to position $j$. Intuitively, $\pi \uplus \pi^*$ can be interpreted as a virtual sequence where both $\pi_j$ and $\pi^*_j$ are added to position $j$ for all $j\in[k]$. Similarly, we define $F(\pi \uplus \pi^*_{\textsf{ovlpd.}})=\sum_{j\in[k]}\lambda_j\cdot f_j(\pi_{[j]}\cup(\pi^*_{[j]}\cap O_{\textsf{ovlpd.}}))$ as the utility of the union of $\pi$ and $O_{\textsf{ovlpd.}}\subseteq \pi^*$. 
Furthermore, we define $F(\pi \uplus \pi^*_{\textsf{cons.}})=\sum_{j\in[k]}\lambda_j\cdot f_j(\pi_{[j]}\cup(\pi^*_{[j]}\cap O_{\textsf{cons.}}))$ as the utility of the union of $\pi$ and  $O_{\textsf{cons.}}\subseteq \pi^*$. 
Finally, we define $F(\pi \uplus \pi^*_{\textsf{not cons.}})=\sum_{j\in[k]}\lambda_j\cdot f_j(\pi_{[j]}\cup(\pi^*_{[j]}\cap O_{\textsf{not cons.}}))$ as the utility of the union of $\pi$ and  $O_{\textsf{not cons.}}\subseteq\pi^*$. 

Throughout this section, all expectations are taken over the coin tosses $\Phi$.  We first provide two technical lemmas.
\begin{lemma}
 \label{lem:3}Let $p\in[0,1]$ denote the sampling probability of each item, $\pi$ denote the output from our algorithm, and $\pi^*$ denote the optimal solution, we have
\begin{eqnarray}
(2+\frac{1}{p})\cdot \mathbb{E} [F(\pi)]\geq  \mathbb{E} [F(\pi \uplus \pi^*)].
\end{eqnarray}
\end{lemma}
\emph{Proof:} By the definition of $O_{\textsf{ovlpd.}}$, all items from $\pi^*_{[j]}\cap O_{\textsf{ovlpd.}}$ must appear in $\pi_{[j]}$ for all $j\in[k]$. Hence, $\pi_{[j]}\cup(\pi^*_{[j]}\cap O_{\textsf{ovlpd.}}) = \pi_{[j]}$. It follows that for all $j\in[k]$, {\small\begin{eqnarray}
&&f_j(\pi_{[j]}\cup(\pi^*_{[j]}\cap O_{\textsf{ovlpd.}})\cup(\pi^*_{[j]}\cap O_{\textsf{not cons.}})\cup(\pi^*_{[j]}\cap O_{\textsf{cons.}})) ~\nonumber\\
&&=f_j(\pi_{[j]}\cup(\pi^*_{[j]}\cap O_{\textsf{not cons.}})\cup(\pi^*_{[j]}\cap O_{\textsf{cons.}})).
\label{eq:1}
 \end{eqnarray}}

Therefore,  $F(\pi \uplus \pi^*)= \sum_{j\in[k]}\lambda_j\cdot f_j(\pi_{[j]}\cup\pi^*_{[j]}) = \sum_{j\in[k]}\lambda_j \cdot f_j(\pi_{[j]}\cup(\pi^*_{[j]}\cap O_{\textsf{ovlpd.}})\cup(\pi^*_{[j]}\cap O_{\textsf{not cons.}})\cup(\pi^*_{[j]}\cap O_{\textsf{cons.}}))=\sum_{j\in[k]}\lambda_j \cdot f_j(\pi_{[j]}\cup(\pi^*_{[j]}\cap O_{\textsf{not cons.}})\cup(\pi^*_{[j]}\cap O_{\textsf{cons.}}))$ where the first equality is by the fact that $O_{\textsf{cons.}}$, $O_{\textsf{not cons.}}$ and $O_{\textsf{ovlpd.}}$ is a partition of $\pi^*$, and the second equality is by equality (\ref{eq:1}).

For simplicity, let  $F(\pi \uplus \pi^*_{\textsf{not cons.}}\uplus \pi^*_{\textsf{cons.}})=\sum_{j\in[k]}\lambda_j\cdot f_j(\pi_{[j]}\cup(\pi^*_{[j]}\cap O_{\textsf{not cons.}})\cup(\pi^*_{[j]}\cap O_{\textsf{cons.}}))$, the above equality indicates that $F(\pi \uplus \pi^*)=F(\pi \uplus \pi^*_{\textsf{not cons.}}\uplus \pi^*_{\textsf{cons.}})$.

 It follows that
\begin{eqnarray}
&& \mathbb{E} [F(\pi \uplus \pi^*)]
= \mathbb{E} [F(\pi)] + \mathbb{E} [F(\pi \uplus \pi^*_{\textsf{not cons.}})- F(\pi)] ~\nonumber\\
&&\quad \quad \quad + \mathbb{E} [F(\pi \uplus \pi^*)- F(\pi \uplus \pi^*_{\textsf{not cons.}})]~\nonumber\\
&&= \mathbb{E} [F(\pi)]+ \mathbb{E} [F(\pi \uplus \pi^*_{\textsf{not cons.}})- F(\pi)] ~\nonumber\\
&& \quad + \mathbb{E} [F(\pi \uplus \pi^*_{\textsf{not cons.}}\uplus \pi^*_{\textsf{cons.}})- F(\pi \uplus \pi^*_{\textsf{not cons.}})]\label{eq:3}
\end{eqnarray}
where the second equality is by the observation that $F(\pi \uplus \pi^*)= F(\pi \uplus \pi^*_{\textsf{not cons.}}\uplus \pi^*_{\textsf{cons.}})$.

Observe that
\begin{eqnarray*}
&&F(\pi \uplus \pi^*_{\textsf{not cons.}}\uplus \pi^*_{\textsf{cons.}})- F(\pi \uplus \pi^*_{\textsf{not cons.}})\\
 &&=  \sum_{j\in[k]}\lambda_j\cdot f_j(\pi_{[j]}\cup(\pi^*_{[j]}\cap O_{\textsf{not cons.}})\cup(\pi^*_{[j]}\cap O_{\textsf{cons.}})) \\
 &&\quad\quad - \sum_{j\in[k]}\lambda_j\cdot f_j(\pi_{[j]}\cup(\pi^*_{[j]}\cap O_{\textsf{not cons.}}))\\
 &&\leq   \sum_{j\in[k]}\lambda_j\cdot f_j(\pi_{[j]}\cup(\pi^*_{[j]}\cap O_{\textsf{cons.}}))- \sum_{j\in[k]}\lambda_j\cdot f_j(\pi_{[j]})\\
 &&= F(\pi \uplus  \pi^*_{\textsf{cons.}})- F(\pi)
\end{eqnarray*}
where the inequality is by the assumption that $f_j$ is submodular for all $j\in [k]$, and the observations that $\pi_{[j]} \subseteq \pi_{[j]}\cup(\pi^*_{[j]}\cap O_{\textsf{not cons.}})$, and $\pi^*_{[j]}\cap O_{\textsf{cons.}}$ does not overlap with $\pi_{[j]}\cup(\pi^*_{[j]}\cap O_{\textsf{not cons.}})$. Here $\pi^*_{[j]}\cap O_{\textsf{cons.}}$ does not overlap with $\pi_{[j]}\cup(\pi^*_{[j]}\cap O_{\textsf{not cons.}})$ is because  $O_{\textsf{cons.}} \cap O_{\textsf{not cons.}} =\emptyset$ (this by the definitions of these two sets) and $O_{\textsf{cons.}}\cap \pi=\emptyset$ (this is due to the fact that any item that is considered but not picked by our algorithm must not appear in $\pi$, noting that each item can be considered only once).

This, together with (\ref{eq:3}), implies that
{\small\begin{eqnarray}
\label{eq:haha1}
&&\mathbb{E} [F(\pi \uplus \pi^*)]
\leq \mathbb{E} [F(\pi)]+ \mathbb{E} [F(\pi \uplus \pi^*_{\textsf{not cons.}})- F(\pi)]~\nonumber\\
&&\quad\quad \quad\quad\quad\quad\quad\quad\quad\quad \quad+ \mathbb{E} [F(\pi \uplus \pi^*_{\textsf{cons.}})- F(\pi)].
\end{eqnarray}}

To prove this lemma, it suffices to show that
$
 p\cdot \mathbb{E} [F(\pi \uplus \pi^*_{\textsf{cons.}})- F(\pi)] \leq \mathbb{E} [F(\pi)]
$
and
\begin{eqnarray}
\label{eq:haha3}
 \mathbb{E} [F(\pi \uplus \pi^*_{\textsf{not cons.}})- F(\pi)]\leq \mathbb{E} [F(\pi)].
\end{eqnarray} The proofs for these two inequalities are provided in the appendix (Section A1) in Lemmas \ref{lem:aaa} and \ref{lem:bbb}, respectively. $\Box$

\begin{lemma}
 \label{lem:4}Let $p\in[0,1]$ denote the sampling probability of each item, $\pi$ denote the output from our algorithm, and $\pi^*$ denote the optimal solution, we have
\begin{eqnarray}
\mathbb{E} [F(\pi \uplus \pi^*)] \geq (1-p) \cdot F(\pi^*).
\end{eqnarray}
\end{lemma}
\emph{Proof:}
We begin by presenting a key result that establishes a link between random sampling and submodular maximization.
\begin{lemma}
\label{lem:2}
(Lemma 2.2 of \cite{buchbinder2014submodular}). Consider a submodular set function $f : 2^V \rightarrow \mathbb{R}$. Let $X$ be a subset of $V$, and let $X(p)$ denote a sampled subset obtained by including each item of $X$ independently with a probability of at most $p$ (not necessarily independent).
The expected value of $f(X(p))$ is at least $(1-p)$ times the value of $f(\emptyset)$.
In other words, $\mathbb{E}[f(X(p))] \geq (1-p)f(\emptyset)$.
\end{lemma}

We define $h_j: 2^V \rightarrow \mathbb{R}$ as follows: $h_j(T) = f_j(T \cup \pi^*_{[j]})$. It can be easily verified that $h_j$ is a submodular function, and $h_j(\emptyset) = f_j(\pi^*_{[j]})$. By applying the above lemma to the function $h_j$ and considering that the items in $\pi_{[j]}$ are chosen with a probability of at most $p$, we can conclude that: $\mathbb{E}[f_j(\pi_{[j]}\cup\pi^*_{[j]})]=\mathbb{E}[h_j(\pi_{[j]})]\geq (1-p)\cdot h_j(\emptyset)=(1-p)\cdot f_j(\pi^*_{[j]})$ for all $j\in [k]$.

The following chain proves this lemma:
\begin{eqnarray}
&&\mathbb{E}[F(\pi \uplus \pi^*)]=\mathbb{E}[\sum_{j\in[k]}\lambda_j\cdot f_j(\pi_{[j]}\cup\pi^*_{[j]})]\\
&&=\sum_{j\in[k]}\lambda_j\cdot \mathbb{E}[f_j(\pi_{[j]}\cup\pi^*_{[j]})]\\
&&\geq \sum_{j\in[k]}\lambda_j\cdot (1-p)\cdot f_j(\pi^*_{[j]})\\
&&=(1-p)\cdot \sum_{j\in[k]}\lambda_j\cdot f_j(\pi^*_{[j]})\\
&&= (1-p) \cdot F(\pi^*).
\end{eqnarray} $\Box$

Lemma \ref{lem:3} and Lemma \ref{lem:4} imply the following main theorem.

\begin{theorem}
\label{thm:1}
Let $p\in[0,1]$ denote the sampling probability of each item, $\pi$ denote the output from our algorithm, and $\pi^*$ denote the optimal solution, we have
\begin{eqnarray}
\mathbb{E} [F(\pi)]\geq \frac{p(1-p)}{2p+1}\cdot F(\pi^*).
\end{eqnarray}
\end{theorem}

It is easy to verify that the maximum value of $\frac{p(1-p)}{2p+1}$ occurs at $p = \frac{\sqrt{3}-1}{2}$, resulting in a maximum value of at least $0.134$.

\begin{corollary}
Let $p=\frac{\sqrt{3}-1}{2}$, we have
\begin{eqnarray}
\mathbb{E} [F(\pi)]\geq 0.134 \cdot F(\pi^*).
\end{eqnarray}
\end{corollary}

\textbf{Remark 1:} For the case when all utility functions $f_j$ are monotone and submodular, Algorithm \ref{alg:1}, by setting $p=1$, can achieve an improved approximation ratio of $1/2$. This is because if $p=1$, then $O_{\textsf{cons.}}=\emptyset$, observing that when $p=1$, all considered items must be added to the solution. It follows that inequality (\ref{eq:haha1}) is reduced to $
\mathbb{E} [F(\pi \uplus \pi^*)] \leq \mathbb{E} [F(\pi)]+ \mathbb{E} [F(\pi \uplus \pi^*_{\textsf{not cons.}})- F(\pi)]$. This can be rewritten as $
\mathbb{E} [F(\pi \uplus \pi^*)] \leq \mathbb{E} [F(\pi \uplus \pi^*_{\textsf{not cons.}})]$. This, together with inequality (\ref{eq:haha3}), implies that $
\mathbb{E} [F(\pi)]\geq  \mathbb{E} [F(\pi \uplus \pi^*)]/2 $. Moreover, if $f_j$ are monotone, it is easy to verify that $\mathbb{E} [F(\pi \uplus \pi^*)]\geq F(\pi^*) $. It follows that $\mathbb{E} [F(\pi)]\geq  \mathbb{E} [F(\pi \uplus \pi^*)]/2  \geq F(\pi^*)/2$.

\textbf{Remark 2:} When all utility functions are homogeneous (i.e., $\exists f: f_j=f$ for all $j\in[k]$), Algorithm \ref{alg:1} becomes independent of the specific values of $\lambda_j$. This is because in Line \ref{line:3} of Algorithm \ref{alg:1}, finding $z$ that maximizes $\sum_{j=t}^{k} \lambda_j\cdot f_j(i\mid \pi)$ is equivalent to maximizing $f(i\mid \pi)$, assuming $f_j=f$ for all $j\in[k]$. Thus, the selection of $z$ becomes independent of $\lambda_j$, showcasing the algorithm's robustness against the knowledge of $\lambda_j$. In the context of recommendation systems, where $\lambda_j$ represents the distribution of customers' patience levels, which may only be estimated approximately or remain unknown, our algorithm consistently delivers high-quality solutions.

\section{NSM with Fixed Length}

We next study the case with fixed length constraints. A formal description of NSM with Fixed Length can be written as follows:
 \begin{center}
\framebox[0.4\textwidth][c]{
\enspace
\begin{minipage}[t]{0.4\textwidth}
\small
$\textbf{P.2}$
$\max_{\pi}  F(\pi)$ subject to $|\pi|= k$.
\end{minipage}
}
\end{center}
\vspace{0.1in}

On the hardness side, achieving a $0.491$-approximation algorithm for $\textbf{P.2}$ is impossible. By setting $\lambda_1=\lambda_2=\cdots, =\lambda_{k-1}=0$ and $\lambda_k=1$ in $\textbf{P.2}$, our problem reduces to maximizing a non-monotone submodular function $f_k$ over a \emph{set} of exactly $k$ items. It is well-known that solving this type of problem does not admit a $0.491$-approximation solution \citep{gharan2011submodular}. Hence, we establish the following lemma.
\begin{lemma}
It is impossible to achieve a $0.491$-approximation for $\textbf{P.2}$.
\end{lemma}

\subsection{Approximation algorithms for $\textbf{P.2}$}
\label{sec:3.2}
The basic idea of our algorithm is to first apply Algorithm \ref{alg:1} to calculate a sequence $\pi$ with a maximum size of $k$, then, if required, we supplement $\pi$ with additional backup items to ensure that it reaches a size of exactly $k$.

We provide a detailed description of our algorithm below:
\begin{enumerate}
 \item Apply  Algorithm \ref{alg:1} to calculate a sequence $\pi$ with a maximum size of $k$.

 \item If  $|\pi|=k$, then return $\pi$ as the final output. Otherwise, randomly select $k - |\pi|$ items $B$ from the set $V\setminus \pi$, and append them to $\pi$ in an arbitrary order. The resulting modified set $\pi$ is then returned as the final output $\pi^{p2}$.
\end{enumerate}

Note that problem $\textbf{P.1}$ is a relaxed problem of $\textbf{P.2}$, observing that any feasible solution of $\textbf{P.2}$ is also a feasible solution of $\textbf{P.1}$. As a consequence, we have the following lemma.
\begin{lemma}
\label{lem:1}
Let $\pi^*$ and $\pi^o$ denote the optimal solution of $\textbf{P.1}$ and $\textbf{P.2}$ respectively, we have
\begin{eqnarray}
F(\pi^*) \geq F(\pi^o).
\end{eqnarray}
\end{lemma}

Now we are ready to present the performance bound of our algorithm.
\begin{theorem}
\label{thm:2}
Let $\pi^{p2}$ denote the solution returned from our algorithm, we have
\begin{eqnarray}
\mathbb{E}_{\Phi, B}[F(\pi^{p2})] \geq (1-\frac{k}{n})\cdot  \frac{p(1-p)}{2p+1}\cdot   F(\pi^o)
\end{eqnarray} where the expectation is taken over the random coin tosses $\Phi$ (phase 1) and the random backup set $B$ (phase 2).
\end{theorem}
\emph{Proof:} To prove this theorem, it suffices to show that
\begin{eqnarray}\label{eq:8}
\mathbb{E}_{\Phi, B}[F(\pi^{p2})] \geq (1-\frac{k}{n})\cdot  \frac{p(1-p)}{2p+1}\cdot F(\pi^*).
\end{eqnarray}
This is because, inequality (\ref{eq:8}), together with Lemma \ref{lem:1}, implies that $\mathbb{E}_{\Phi, B}[F(\pi^{p2})] \geq (1-\frac{k}{n})\cdot  \frac{p(1-p)}{2p+1}\cdot F(\pi^*) \geq  (1-\frac{k}{n})\cdot  \frac{p(1-p)}{2p+1}\cdot F(\pi^o)$.

The remaining part of the proof focuses on establishing inequality (\ref{eq:8}). Recall that our algorithm first use Algorithm \ref{alg:1} to compute a sequence $\pi$. Consider a specific run of Algorithm \ref{alg:1} and its resulting sequence $\pi$. Recall that $\pi^{p2}$ is constructed by appending $B$ to $\pi$, where $B$ is a random set of backup items from $V\setminus \pi$ such that $|B| = k - |\pi|$. Hence, we can express the utility of $\pi^{p2}$ as follows:
\begin{eqnarray}
F(\pi^{p2}) = \sum_{j \in [k]} \lambda_j \cdot f_j(\pi_{[j]} \cup (\pi^{p2}_{[j]} \cap B)).
\end{eqnarray}
Here,  $\pi^{p2}_{[j]}$ represents the first $j$ items of $\pi^{p2}$. Because $B$, with a size of $k - |\pi|$, is a subset randomly chosen from the set $V\setminus \pi$, $\pi^{p2}_{[j]} \cap B$ represents a randomly selected set from $V\setminus \pi$, with a size of at most $k - |\pi|$. Consequently, the probability that each item from $V\setminus \pi$ is added to $\pi^{p2}_{[j]} \cap B$ is at most $\frac{k - |\pi|}{n-|\pi|}$, where $n-|\pi|$ represents the size of $V\setminus \pi$.  Note that  $\frac{k - |\pi|}{n-|\pi|} \leq k/n$.

By abuse of notation, define $h_j: 2^V \rightarrow \mathbb{R}$ as follows: $h_j(T) = f_j(T \cup \pi_{[j]})$. By applying Lemma \ref{lem:2} to the function $h_j$ and considering that the items in $\pi^{p2}_{[j]} \cap B$ are chosen with a probability of at most $k/n$, we can conclude that: \begin{eqnarray}&&\mathbb{E}_B[f_j(\pi_{[j]}\cup(\pi^{p2}_{[j]} \cap B) )]=\mathbb{E}_B[h_j(\pi^{p2}_{[j]} \cap B)] \\
&&\geq (1-k/n)\cdot h_j(\emptyset)=(1-k/n)\cdot f_j(\pi_{[j]})\label{eq:9}\end{eqnarray}
for all $j\in [k]$.

For a given set of coin tosses $\Phi$ and the resulting output sequence $\pi$ from phase 1, we can conclude that
\begin{eqnarray}
&&\mathbb{E}_B[F(\pi^{p2})] = \mathbb{E}_B[\sum_{j \in [k]} \lambda_j \cdot f_j(\pi_{[j]} \cup (\pi^{p2}_{[j]} \cap B))]\\
&&= \sum_{j \in [k]} \lambda_j \cdot \mathbb{E}_B[ f_j(\pi_{[j]} \cup (\pi^{p2}_{[j]} \cap B))]\\
&&\geq \sum_{j \in [k]} \lambda_j \cdot (1-k/n)\cdot f_j(\pi_{[j]})\\
&&=  (1-k/n)\cdot \sum_{j \in [k]} \lambda_j \cdot f_j(\pi_{[j]}) = (1-k/n)\cdot  F(\pi)
\end{eqnarray}
where the inequality is by inequality (\ref{eq:9}).

Taking the expectation of both sizes of the above inequality over the random coin tosses $\Phi$, we have
\begin{eqnarray}
\mathbb{E}_{\Phi, B}[F(\pi^{p2})]\geq \mathbb{E}_{\Phi}[(1-k/n)\cdot  F(\pi)]= (1-k/n)\cdot \mathbb{E}_{\Phi}[F(\pi)].
\end{eqnarray}

This, together with   Theorem \ref{thm:1}, implies that
\begin{eqnarray}
\mathbb{E}_{\Phi, B}[F(\pi^{p2})]\geq  (1-k/n)\cdot \mathbb{E}_{\Phi}[F(\pi)] \geq (1-\frac{k}{n})\cdot  \frac{p(1-p)}{2p+1}\cdot   F(\pi^o).
\end{eqnarray}

This finishes the proof of inequality (\ref{eq:8}). As a result, the theorem has been proven. $\Box$

If we set $p=\frac{\sqrt{3}-1}{2}$ in the first phase, we have the following corollary.

\begin{corollary}
\label{cor:1}
Let $p=\frac{\sqrt{3}-1}{2}$, we have
\begin{eqnarray}
\mathbb{E}_{\Phi, B}[F(\pi^{p2})]\geq (1-\frac{k}{n})\cdot  0.134 \cdot   F(\pi^o).
\end{eqnarray}
\end{corollary}

The above corollary implies that when the size constraint $k$ remains relatively small compared to the overall number of items $n$, our algorithm is capable of achieving a good approximation of the optimal result. This observation is particularly relevant in scenarios such as recommendation systems where $k$ is often much smaller than $n$.
\subsection{Constant-factor approximation algorithm for homogeneous functions}

Next, we will examine an important special case of $\textbf{P.2}$, where we assume that all utility functions are homogeneous. This means that there exists a submodular function $f: 2^V \rightarrow \mathbb{R}^+$ such that $f_j = f$ for all $j \in [k]$. A formal description of this special case is listed in below.

 \begin{center}
\framebox[0.6\textwidth][c]{
\enspace
\begin{minipage}[t]{0.6\textwidth}
\small
$\textbf{P.2}$
$\max_{\pi}  \sum_{j\in[k]}\lambda_j\cdot f(\pi_{[j]})$ subject to $|\pi|= k$.
\end{minipage}
}
\end{center}
\vspace{0.1in}

Next, we present constant-factor approximation algorithms for this special case, which represents an improvement over the general case where we can only achieve an approximation ratio based on $k/n$.

We consider two cases: when $k< n/2$ and when $k\geq n/2$. The case when $k< n/2$ is straightforward, as one can directly apply our algorithm developed for the general case (as presented in the previous section) to achieve a constant-factor approximation. This is because when $k< n/2$, we have $1-k/n>1/2$. This, together with Corollary \ref{cor:1}, implies the following corollary.

\begin{corollary}
\label{cor:11}
Assume $k< n/2$, let $p=\frac{\sqrt{3}-1}{2}$, we have
\begin{eqnarray}
\mathbb{E}_{\Phi, B}[F(\pi^{p2})]\geq \frac{0.134}{2} \cdot   F(\pi^o)
\end{eqnarray} where $\pi^o$ is the optimal solution of $\textbf{P.2}$.
\end{corollary}

The rest of this section focuses on the case when $k\geq n/2$. For the sake of simplicity, let's assume that $n$ is an even number. Considering the optimal solution $\pi^o$, we can partition its utility $F(\pi^o) $ into two parts:

\begin{equation}
F(\pi^o) = \sum_{j \in [\frac{n}{2}]} \lambda_j \cdot f(\pi^o_{[j]}) + \sum_{j \in \{\frac{n}{2}+1, \cdots, k\}} \lambda_j \cdot f(\pi^o_{[j]}).
\end{equation}

Here, the first part represents the utility obtained from the first $n/2$ functions, while the second part represents the utility obtained from the remaining $k - n/2$ functions. We can further divide the analysis into two subcases based on the relationship between the utilities from these two parts. Intuitively, if the first part dominates the overall utility, it suffices to find a sequence that maximizes the utility from the first part. Otherwise, if the second part contributes significantly to the overall utility, our focus is on finding a solution maximizes the second part. Although we initially have no information about the relationship between the two parts of the utility, given that $\pi^o$ is unknown, we can make a guess regarding this relation and consider both possibilities. We can evaluate the performance of each case and choose the one that yields the best overall utility as the final output.

\subsubsection{The case when $\sum_{j \in [\frac{n}{2}]} \lambda_j \cdot f(\pi^o_{[j]}) \geq \sum_{j \in \{\frac{n}{2}+1, \cdots, k\}} \lambda_j \cdot f(\pi^o_{[j]})$}
\label{sec:1}
We first study the case when $\sum_{j \in [\frac{n}{2}]} \lambda_j \cdot f(\pi^o_{[j]}) \geq \sum_{j \in \{\frac{n}{2}+1, \cdots, k\}} \lambda_j \cdot f(\pi^o_{[j]})$, that is, the first $n/2$ functions contributes more than half of the total utility. In this scenario, our objective is to find a sequence that maximizes the utility from the first part (a formal description of this problem is listed in $\textbf{P.2a}$).

 \begin{center}
\framebox[0.6\textwidth][c]{
\enspace
\begin{minipage}[t]{0.6\textwidth}
\small
$\textbf{P.2a}$
$\max_{\pi}  \sum_{j\in[\frac{n}{2}]}\lambda_j\cdot f(\pi_{[j]})$ subject to $|\pi|= k$.
\end{minipage}
}
\end{center}
\vspace{0.1in}

Let $\pi^a$ denote the optimal solution of $\textbf{P.2a}$, we have
\begin{eqnarray}
\label{eq:10}
\sum_{j \in [\frac{n}{2}]} \lambda_j \cdot f(\pi^a_{[j]}) \geq \sum_{j \in [\frac{n}{2}]} \lambda_j \cdot f(\pi^o_{[j]}) \geq \frac{1}{2}\cdot F(\pi^o)
\end{eqnarray}
where the first inequality is because $\pi^o$ is a feasible solution of $\textbf{P.2a}$ and the assumption that $\pi^a$ is an optimal solution of $\textbf{P.2a}$, and the second inequality is by the fact that $F(\pi^o) = \sum_{j \in [\frac{n}{2}]} \lambda_j \cdot f(\pi^o_{[j]}) + \sum_{j \in \{\frac{n}{2}+1, \cdots, k\}} \lambda_j \cdot f(\pi^o_{[j]})$ and the assumption that  $\sum_{j \in [\frac{n}{2}]} \lambda_j \cdot f(\pi^o_{[j]}) \geq \sum_{j \in \{\frac{n}{2}+1, \cdots, k\}} \lambda_j \cdot f(\pi^o_{[j]})$.

Now we are ready to present our algorithm. It is easy to verify that if we replace the constraint $|\pi|= k$ in $\textbf{P.2a}$ by $|\pi|= \frac{n}{2}$, it does not affect the value of the optimal solution. This is because those items placed after position $n/2$ do not affect the utility of the first $n/2$ functions anyway. A formal description of this variant is listed in $\textbf{P.2b}$.

 \begin{center}
\framebox[0.6\textwidth][c]{
\enspace
\begin{minipage}[t]{0.6\textwidth}
\small
$\textbf{P.2b}$
$\max_{\pi}  \sum_{j\in[\frac{n}{2}]}\lambda_j\cdot f(\pi_{[j]})$ subject to $|\pi|= \frac{n}{2}$.
\end{minipage}
}
\end{center}
\vspace{0.1in}

Let $\pi^b$ denote the optimal solution of $\textbf{P.2b}$, we have
\begin{eqnarray}
\label{eq:hawai}
\sum_{j \in [\frac{n}{2}]} \lambda_j \cdot f(\pi^b_{[j]}) = \sum_{j \in [\frac{n}{2}]} \lambda_j \cdot f(\pi^a_{[j]}) \geq \frac{1}{2}\cdot F(\pi^o)
\end{eqnarray}
where the equality is because, as previously discussed, $\textbf{P.2a}$ and  $\textbf{P.2b}$ share the same value of the optimal solution, and the  inequality is by inequality (\ref{eq:10}).

We can utilize our algorithm designed for the general case to solve $\textbf{P.2b}$ and derive a solution denoted as $\pi$. Considering that the size constraint in $\textbf{P.2b}$ is $\frac{n}{2}$, we can substitute $\frac{n}{2}$ into Corollary \ref{cor:1} and conclude that $\pi$ provides a $\frac{0.134}{2}$-approximation solution for $\textbf{P.2b}$, that is,
\begin{eqnarray}\label{eq:11}
\sum_{j \in [\frac{n}{2}]} \lambda_j \cdot f(\pi_{[j]}) \geq \frac{0.134}{2}\cdot \sum_{j \in [\frac{n}{2}]} \lambda_j \cdot f(\pi^b_{[j]}).
\end{eqnarray}
However,  $\pi$ might not be a feasible solution for the original problem $\textbf{P.2a}$, as its size might be less than $k$.
To ensure feasibility, we can append an arbitrary set of $k - \frac{n}{2}$ items from $V\setminus \pi$ to $\pi$ to obtain the final solution $\pi'$.  This step ensures that the solution satisfies the size constraint and makes it feasible for $\textbf{P.2a}$. More importantly, this step does not affect the utility of $\pi$, given that any items placed after position $n/2$ do not affect the utility of the first $n/2$ functions. As a result, we achieve a $\frac{0.134}{4}$-approximation for $\textbf{P.2}$, that is, $F(\pi')\geq \sum_{j \in [\frac{n}{2}]} \lambda_j \cdot f(\pi'_{[j]})= \sum_{j \in [\frac{n}{2}]} \lambda_j \cdot f(\pi_{[j]}) \geq \frac{0.134}{2}\cdot \sum_{j \in [\frac{n}{2}]} \lambda_j \cdot f(\pi^b_{[j]}) \geq \frac{0.134}{4}\cdot F(\pi^o)$ where the equality is because, as previously discussed,  any items placed after position $n/2$ do not affect the utility of the first $n/2$ functions, the second inequality by inequality (\ref{eq:11}) and the third inequality is by inequality (\ref{eq:hawai}). This leads to the following lemma.

\begin{lemma}
There exists a $\frac{0.134}{4}$-approximation solution for $\textbf{P.2}$, assuming  $\sum_{j \in [\frac{n}{2}]} \lambda_j \cdot f(\pi^o_{[j]}) \geq \sum_{j \in \{\frac{n}{2}+1, \cdots, k\}} \lambda_j \cdot f(\pi^o_{[j]})$.
\end{lemma}

\subsubsection{The case when $\sum_{j \in [\frac{n}{2}]} \lambda_j \cdot f(\pi^o_{[j]}) < \sum_{j \in \{\frac{n}{2}+1, \cdots, k\}} \lambda_j \cdot f(\pi^o_{[j]})$}
\label{sec:2}
We next study the case when $\sum_{j \in [\frac{n}{2}]} \lambda_j \cdot f(\pi^o_{[j]}) < \sum_{j \in \{\frac{n}{2}+1, \cdots, k\}} \lambda_j \cdot f(\pi^o_{[j]})$, that is, the latter half of the functions contributes more to the overall utility than the first half. In this scenario, our objective is to find a sequence that maximizes the utility from the latter half of the functions (a formal description of this problem is listed in $\textbf{P.2c}$).

 \begin{center}
\framebox[0.6\textwidth][c]{
\enspace
\begin{minipage}[t]{0.6\textwidth}
\small
$\textbf{P.2c}$
$\max_{\pi}  \sum_{j\in \{\frac{n}{2}+1, \cdots, k\} }\lambda_j\cdot f(\pi_{[j]})$ subject to $|\pi|= k$.
\end{minipage}
}
\end{center}
\vspace{0.1in}

Let $\pi^c$ denote the optimal solution of $\textbf{P.2c}$, we have
\begin{eqnarray}
\label{eq:101}
\sum_{j \in \{\frac{n}{2}+1, \cdots, k\}} \lambda_j \cdot f(\pi^c_{[j]}) \geq \sum_{j \in \{\frac{n}{2}+1, \cdots, k\}} \lambda_j \cdot f(\pi^o_{[j]}) \geq \frac{1}{2}\cdot F(\pi^o)
\end{eqnarray}
where the first inequality is by the observation that $\pi^o$ is a feasible solution of $\textbf{P.2c}$ and the assumption that $\pi^c$ is the optimal solution of $\textbf{P.2c}$,  and the second inequality is by the assumption that $\sum_{j \in [\frac{n}{2}]} \lambda_j \cdot f(\pi^o_{[j]}) < \sum_{j \in \{\frac{n}{2}+1, \cdots, k\}} \lambda_j \cdot f(\pi^o_{[j]})$.

Before presenting our algorithm, we first introduce a new utility function $g: 2^V\rightarrow \mathbb{R}_+$ as below:
\begin{eqnarray}
g(\cdot) = f(V\setminus \cdot).
\end{eqnarray}

We next introduce a well-known result that states the submodularity property of functions is preserved when considering their complement.
\begin{lemma}
\label{lem:submodular}
If $f$ is submodular, then $g$ must be submodular.
\end{lemma}

Now we are ready to present our algorithm which is a sampling-based greedy algorithm with respect to function $g$. Our algorithm consists of three phases:
\begin{enumerate}
 \item We begin by selecting a random subset, denoted as $R$, from the set of items $V$. Each item $i \in V$ is included in $R$ independently with a probability $p \in [0, 1]$, where $p$ is a parameter that will be optimized later.
 \item  We employ a greedy algorithm  on $R$. The greedy algorithm operates by iteratively adding items to the current solution based on their marginal utility with respect to the function $g$.  To be precise, let $U$, whose initial value is $\emptyset$, denote the current solution. Let $z\in \argmax_{i\in R} g(i\mid U)$ denote the item that has the largest marginal utility  on top of $U$ w.r.t. function $g$. If $g(z\mid U)>0$, then add $z$ to $U$ (i.e., $U = U\cup\{z\}$).  The construction process continues until one of the two stopping criteria is met: either the solution reaches a size of $n/2$, or the marginal utility of the remaining items becomes non-positive.  Let $U$ denote the solution returned from this phase.
 \item Finally, we randomly select a set of $n/2-|U|$ items $B$ from $V\setminus U$ as a backup set. Assume $\pi^U$ denotes the sequence of the last $|U|-n+k$ items added to $U$, in reverse order based on the time they were added to $U$. For the case when $|U|-n+k\leq 0$, we set $\pi^U=\emptyset$.  For example, suppose the last $|U|-n+k$ items added to $U$ are ${3, 5, 9}$ in accordance with the order of their addition to $U$, then its corresponding $\pi^U$ is $\{9, 5, 3\}$. Let $\pi^B$ denote the sequence of items in the backup set $B$ arranged in an arbitrary order, and let $\pi^{V\setminus(U\cup B)}$ denote the sequence of the items in $V\setminus(U\cup B)$ arranged in an  arbitrary order. The final output of our algorithm is $\pi^{V\setminus(U\cup B)}\oplus \pi^B\oplus \pi^U$.
\end{enumerate}

\begin{algorithm}[hptb]
\caption{\textsf{Sampling-Greedy w.r.t. function $g$}}
\label{alg:2}
\begin{algorithmic}[1]
\STATE $E=V$, $U=\emptyset$, $\pi^U =\emptyset$, $\pi^B =\emptyset$, $\pi^{V\setminus(U\cup B)} =\emptyset$, $Q=\{i\in V\mid g(\{i\})>0\}$, $t=1$
\WHILE {$t\leq n/2$ and $Q\neq \emptyset$}
\STATE \underline{consider} $z=\argmax_{i\in Q}  g(i\mid U)$
\STATE $E=E\setminus \{z\}$
\STATE let $\Phi_z \sim \mathrm{Bernoulli}(p)$
\IF {$\Phi_z=1$}
\STATE $U = U\cup\{z\}$
\IF {$t \geq n-k$}
\STATE $\pi^U = z \oplus \pi^U$
\ENDIF
\STATE $t\leftarrow t+1$,  $Q=\{i\in E\mid g(i \mid U)>0\}$
\ENDIF
\ENDWHILE
\STATE randomly select a subset $B$ of $\frac{n}{2}-|U|$ items from the set $V\setminus U$
\STATE let $\pi^B$ represent an arbitrary sequence of the items in $B$
\STATE let $\pi^{V\setminus(U\cup B)}$ denote an arbitrary sequence of items from the set $V\setminus(U\cup B)$
\STATE $\pi''\leftarrow \pi^{V\setminus(U\cup B)}\oplus \pi^B\oplus \pi^U$
\RETURN $\pi''$
\end{algorithmic}
\end{algorithm}

 Once again, with the purpose of analysis, we introduce an alternative implementation of the aforementioned algorithm. Specifically, after evaluating each item, we introduce a coin toss with a success probability of $p$ to determine whether the item should be included in the solution.  A detailed description of this alternative is listed in Algorithm \ref{alg:2}.

\textbf{A running example of the algorithm.} Suppose we have a set of items $V = \{1, 2, 3, 4, 5, 6, 7, 8, 9, 10\}$, and we want to select a sequence of $k=8$ items. We randomly select a subset $R$ from $V$, using  $p = 0.6$. Let's say $R = \{1, 4, 5, 7, 8, 9, 10\}$. Then we apply a greedy algorithm on $R$ to iteratively select items based on their marginal utility with respect to function $g$.  Let's assume that the greedy algorithm terminates after adding items $5$, $1$, $9$, and $10$ to $U$. So, $U = \{5, 1, 9, 10\}$. We define the sequence of the last $|U| - n + k = 4 - 10 + 8 = 2$ items added to $U$ in reverse order as $\pi^U$.  Thus, $\pi^U = \{10, 9\}$.  We randomly select a backup set $B$ of $n/2 - |U| = 10/2 - 4 = 1$ items from $V\setminus U$. Let's say $B = \{2\}$.  We arrange the items in the backup set $B$ in an arbitrary order and denote it as $\pi^B$. Since $B$ is a singleton, $\pi^B = \{2\}$.  We arrange the items in $V\setminus(U\cup B)$ in an arbitrary order and denote it as $\pi^{V\setminus(U\cup B)}$.  Since $V\setminus(U\cup B) = \{3, 4, 6, 7, 8\}$, let's assume $\pi^{V\setminus(U\cup B)} = \{3, 4, 6, 7, 8\}$. The final output of the algorithm is $\pi^{V\setminus(U\cup B)}\oplus \pi^B\oplus \pi^U$. Plugging in the values, the final output sequence is $\{3, 4, 6, 7, 8, 2, 10, 9\}$.

\textbf{Performance analysis.}
To analyze the performance of our algorithm, let us consider the following optimization problem:

 \begin{center}
\framebox[0.6\textwidth][c]{
\enspace
\begin{minipage}[t]{0.6\textwidth}
\small
$\textbf{P.2d}(j)$
$\max_{\pi}  \lambda_j\cdot f(\pi_{[j]})$ subject to $|\pi|= j$.
\end{minipage}
}
\end{center}
\vspace{0.1in}

For a given $j\in \{\frac{n}{2}+1, \cdots, k\}$, $\textbf{P.2d}(j)$ aims to find a best sequence, whose size is $j$, that maximizes the $j$-th function.  It is easy to verify that the ordering of the selected items does not impact the resulting utility. Solving $\textbf{P.2d}(j)$ is equivalent to solving the following problem $\textbf{P.2e}(j)$, a flipped version of $\textbf{P.2d}(j)$:

 \begin{center}
\framebox[0.6\textwidth][c]{
\enspace
\begin{minipage}[t]{0.6\textwidth}
\small
$\textbf{P.2e}(j)$
$\max_{S\subseteq V}  \lambda_j\cdot g(S)$ subject to $|S|= n-j$.
\end{minipage}
}
\end{center}
\vspace{0.1in}

To solve $\textbf{P.2e}(j)$, and hence $\textbf{P.2d}(j)$, for any $j\in \{\frac{n}{2}+1, \cdots, k\}$, we present a sampling-based greedy algorithm (labeled as \textsf{Sampling-Greedy-$j$}), which consists of three phases:
\begin{enumerate}
 \item We begin by selecting a random subset, denoted as $R$, from the set of items $V$. Each item $i \in V$ is included in $R$ independently with a probability $p \in [0, 1]$, where $p$ is a parameter that will be optimized later.

 \item  We employ a greedy algorithm  on $R$. The greedy algorithm operates by iteratively adding items to the current solution based on their marginal utility with respect to the function $g$. The construction process continues until one of the two stopping criteria is met: either the solution reaches a size of $n-j$, or the marginal utility of the remaining items becomes non-positive.  Let $S$ denote the solution returned from this phase.
 \item To ensure that the solution reaches a size of $n-j$, we randomly select a set of $n-j-|S|$ items $B$ from $V\setminus S$ as a backup set and add them to $S$. Let $S^{j}= S \cup B$ denote the final output.
\end{enumerate}

To assess the performance of \textsf{Sampling-Greedy-$j$}, we focus on an alternative implementation of the algorithm. In this alternative approach, similar to Algorithm \ref{alg:1} and \ref{alg:2}, we postpone the selection of $R$ by introducing a coin toss with a success probability of $p$ to determine whether an item should be included in the solution.

\begin{theorem}
Let $O^j$ denote the optimal solution of $\textbf{P.2e}(j)$ and $S^j$ denote the output from \textsf{Sampling-Greedy-$j$},  for any $j\in \{\frac{n}{2}+1, \cdots, k\}$, we have
\begin{eqnarray}
\mathbb{E}_{\Phi, B}[g(S^{j})] \geq \frac{1}{2}\cdot  \frac{p(1-p)}{2p+1}\cdot   g(O^j)
\end{eqnarray} where the expectation is taken over the random coin tosses $\Phi$  and the random backup set $B$.
\end{theorem}
\emph{Proof:} Observing that our algorithm is identical to the one as introduced in Section \ref{sec:3.2}, moreover, the function $g$ is submodular,  by following the same proof of Theorem \ref{thm:2}, we can show that
\begin{eqnarray}\label{eq:12}\mathbb{E}_{\Phi, B}[g(S^{j})]\geq (1-q)\cdot  \frac{p(1-p)}{2p+1}\cdot   g(O^j)
 \end{eqnarray} where $q$ is the largest possible probability that each item from $V\setminus S$ is being added to the backup set $B$. It is easy to verify that the probability that each item from $V\setminus S$ is being added to  $B$ is $(n-j-|S|)/(n-|S|)$. It follows that $q\leq (n-j-|S|)/(n-|S|)\leq (n-j)/n \leq 1/2$ where the second inequality is by the assumption that $j\in \{\frac{n}{2}+1, \cdots, k\}$. This, together with inequality (\ref{eq:12}), implies that $\mathbb{E}_{\Phi, B}[g(S^{j})]\geq (1/2)\cdot  \frac{p(1-p)}{2p+1}\cdot   g(O^j)$. $\Box$

From the above theorem and the fact that $\textbf{P.2d}(j)$ and $\textbf{P.2e}(j)$ share the same value of the optimal solution, the following corollary can be readily inferred:
\begin{corollary}
Let $\pi^{e(j)}$ denote the optimal solution of $\textbf{P.2d}(j)$ and $S^j$ denote the solution returned from \textsf{Sampling-Greedy-$j$},  for any $j\in \{\frac{n}{2}+1, \cdots, k\}$, we have
\begin{eqnarray}
\label{eq:13}
\mathbb{E}_{\Phi, B}[f(V\setminus S^{j})] \geq \frac{1}{2}\cdot  \frac{p(1-p)}{2p+1}\cdot   f(\pi^{e(j)}).
\end{eqnarray}
\end{corollary}

Now we are ready to present the main theorem which states that Algorithm \ref{alg:2} achieves an approximation ratio of $\frac{1}{4}\cdot  \frac{p(1-p)}{2p+1}$.

\begin{theorem}Let $\pi''$ denote the solution returned from Algorithm \ref{alg:2} and $\pi^o$ denote the optimal solution of $\textbf{P.2}$, assuming $\sum_{j \in [\frac{n}{2}]} \lambda_j \cdot f(\pi^o_{[j]}) < \sum_{j \in \{\frac{n}{2}+1, \cdots, k\}} \lambda_j \cdot f(\pi^o_{[j]})$, we have
   \begin{eqnarray}\label{eq:12d}
\mathbb{E}[ F(\pi'')] \geq \frac{1}{4}\cdot  \frac{p(1-p)}{2p+1}\cdot F(\pi^o).
 \end{eqnarray}
\end{theorem}
\emph{Proof:} Recall that $\pi^c$ denotes the optimal solution of $\textbf{P.2c}$, we have
\begin{eqnarray}\label{eq:121}
\sum_{j \in \{\frac{n}{2}+1, \cdots, k\}} \lambda_j \cdot f(\pi^c_{[j]}) \leq \sum_{j \in \{\frac{n}{2}+1, \cdots, k\}} \lambda_j \cdot f(\pi^{e(j)})
 \end{eqnarray}
 where the inequality is by the observation that for all $j \in \{\frac{n}{2}+1, \cdots, k\}$, $\pi^c_{[j]}$ is a feasible solution of $\textbf{P.2d}(j)$ and the assumption that $\pi^{e(j)}$ is the optimal solution of $\textbf{P.2d}(j)$. This, together with inequality (\ref{eq:13}), implies that
 \begin{eqnarray}\label{eq:121}
\frac{1}{2}\cdot  \frac{p(1-p)}{2p+1}\cdot\sum_{j \in \{\frac{n}{2}+1, \cdots, k\}} \lambda_j \cdot f(\pi^c_{[j]})  \leq \sum_{j \in \{\frac{n}{2}+1, \cdots, k\}} \lambda_j \cdot \mathbb{E}_{\Phi, B}[f(V\setminus S^{j})].
 \end{eqnarray}

 This, together with inequality (\ref{eq:101}), implies that
  \begin{eqnarray}\label{eq:1211}
\frac{1}{2}\times \frac{1}{2}\cdot  \frac{p(1-p)}{2p+1}\cdot F(\pi^o) \leq \sum_{j \in \{\frac{n}{2}+1, \cdots, k\}} \lambda_j \cdot \mathbb{E}_{\Phi, B}[f(V\setminus S^{j})].
 \end{eqnarray}

 To prove this theorem, it suffices to show that
   \begin{eqnarray}\label{eq:12112}
\mathbb{E}[ F(\pi'')] \geq \sum_{j \in \{\frac{n}{2}+1, \cdots, k\}} \lambda_j \cdot \mathbb{E}_{\Phi, B}[f(V\setminus S^{j})].
 \end{eqnarray}

The remainder of the proof focuses on establishing the above equality. Let's recall that $U$ represents the set returned from the second phase of Algorithm \ref{alg:2}, obtained using a greedy algorithm. For every $j \in \{\frac{n}{2}+1, \cdots, k\}$, let $U^j$ denote the first $\min\{n-j, |U|\}$ items that are added to $U$. It is important to emphasize that for every $j \in \{\frac{n}{2}+1, \cdots, k\}$, the distributions of the sets $S$ (i.e., the set returned from the second phase of \textsf{Sampling-Greedy-$j$}) and $U^j$ are identical. Furthermore, it is worth noting that the backup set, consisting of $\max\{n-j-|U|, 0\}$ items, added to $S$, and the last $\max\{n-j-|U|, 0\}$ items of $\pi^B$ in Algorithm \ref{alg:2}, also exhibit the same distribution. As a result, for every $j \in \{\frac{n}{2}+1, \cdots, k\}$, $\pi''_{[j]}$ share the same distribution of $V\setminus S^{j}$. It follows that
   \begin{eqnarray}\label{eq:12111}
\mathbb{E}[ F(\pi'')] = \sum_{j \in [k]} \lambda_j \cdot \mathbb{E}[f(\pi''_{[j]})] \geq \sum_{j \in \{\frac{n}{2}+1, \cdots, k\}} \lambda_j \cdot \mathbb{E}[f(\pi''_{[j]})] = \sum_{j \in \{\frac{n}{2}+1, \cdots, k\}} \lambda_j \cdot \mathbb{E}_{\Phi, B}[f(V\setminus S^{j})].
 \end{eqnarray}
This finishes the proof of inequality (\ref{eq:12112}). $\Box$

The following corollary follows immediately from the above theorem by setting $p=\frac{\sqrt{3}-1}{2}$:
\begin{corollary}
There exists a $\frac{0.134}{4}$-approximation solution for $\textbf{P.2}$, assuming  $\sum_{j \in [\frac{n}{2}]} \lambda_j \cdot f(\pi^o_{[j]}) < \sum_{j \in \{\frac{n}{2}+1, \cdots, k\}} \lambda_j \cdot f(\pi^o_{[j]})$.
\end{corollary}

\subsubsection{Putting it all together}
Recall that we develop $\frac{0.134}{4}$-approximation algorithms for the case when $\sum_{j \in [\frac{n}{2}]} \lambda_j \cdot f(\pi^o_{[j]}) \geq \sum_{j \in \{\frac{n}{2}+1, \cdots, k\}} \lambda_j \cdot f(\pi^o_{[j]})$ and $\sum_{j \in [\frac{n}{2}]} \lambda_j \cdot f(\pi^o_{[j]}) < \sum_{j \in \{\frac{n}{2}+1, \cdots, k\}} \lambda_j \cdot f(\pi^o_{[j]})$ respectively. Although we lack prior knowledge of the optimal solution, selecting the superior solution between the aforementioned options guarantees achieving an approximation ratio of $\frac{0.134}{4}$.

\section{Experimental Evaluation}

We conduct experiments on real-world datasets to evaluate the impact of user type distributions in the context of video recommendation. Suppose we are dealing with a large video library, denoted as $V$, containing videos spanning multiple categories, each categorized as potentially overlapping subsets, namely $C_1, C_2, \ldots, C_m \subseteq V$. When a user provides a set of category preferences, our platform's objective is to generate a sequence of videos, denoted as $\pi$, from those specified categories that maximizes the expected user engagement $\sum_{j\in[k]}\lambda_j\cdot f_j(\pi_{[j]})$. Recall that $k$ denotes the maximum window size of displayed videos, and $\lambda_j$ represents the proportion of users with a specific patience level $j$ who are willing to view the first $j$ videos $\pi_{[j]}$. Denote by $\mathcal{U}$ the space of user types, each user type $u\in\mathcal{U}$ is specified by a pair $(j, f_j(\cdot))$. The user considers the first $j$ videos in the list and obtains utility $f_j(\pi_{[j]})$.
The platform lacks precise knowledge of the user's exact type but is aware of the type distribution $\mathcal{D}$.

We consider a common characterization of $f_j(\cdot)$ as a submodular function. Each video $s$ has a rating $\rho_s$ and we denote by $w_{st}\in[0, 1]$ some measure of the percentage of similarity between any two videos $s$ and $t$.
In introducing our function $f_j(\cdot)$, we adopt the approach outlined in \citep{amanatidis2020fast}. We start by considering the auxiliary objective $g_j(\pi_{[j]})=\sum_{s\in\pi_{[j]}}\sum_{t\in V}w_{st}-\eta\sum_{s\in\pi_{[j]}}\sum_{t\in\pi_{[j]}}w_{st}$, for some $\eta\geq1$ \citep{mirzasoleiman2016fast}. This objective takes inspiration from maximal marginal relevance \citep{carbonell1998use}, emphasizing coverage while penalizing similarity.  To strike a balance between selecting highly-rated videos and those that comprehensively represent the entire collection, we employ the submodular function  $f_j(\pi_{[j]})=\alpha\sum_{s\in\pi_{[j]}}\rho_s+\beta g_j(\pi_{[j]})$ for $\alpha,\beta\geq0$.

\textbf{Datasets.} We assess our proposed algorithms and benchmarks using the latest MovieLens dataset \citep{harper2015movielens}, which comprises $62,423$ movies, with $13,816$ having both user-generated tags and ratings. Similarities, denoted as $w_{st}$, are computed from these tags using the L2 norm of the pairwise minimum tag vector, with additional details to follow.

\begin{figure*}[hpt]
\begin{center}
\includegraphics[scale=0.3]{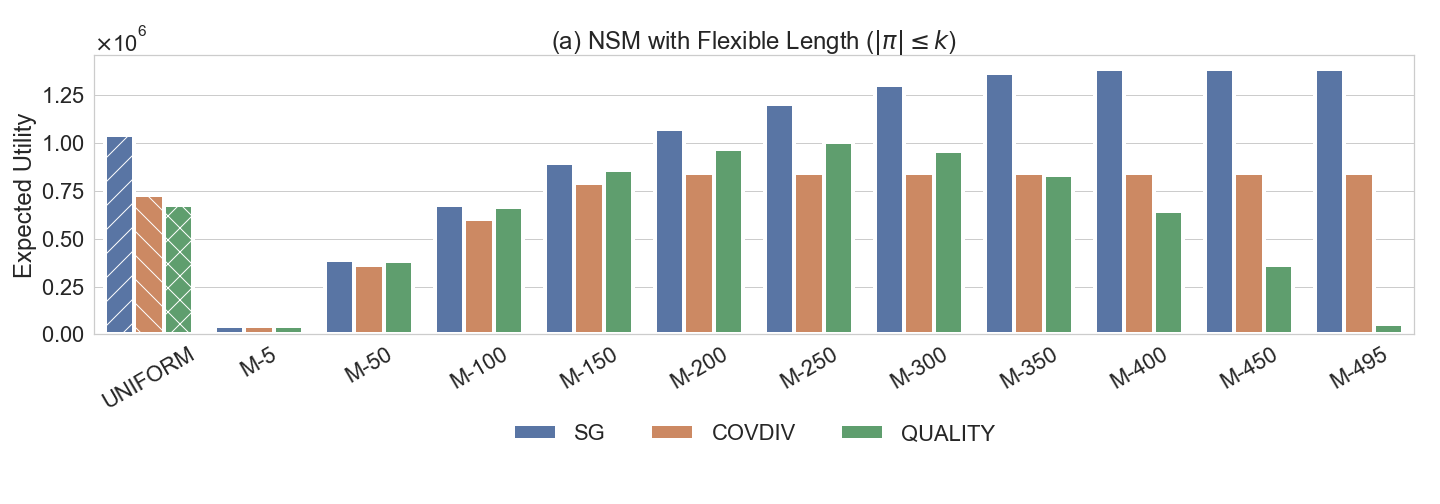}\\
\vspace*{-0.2in}
\includegraphics[scale=0.3]{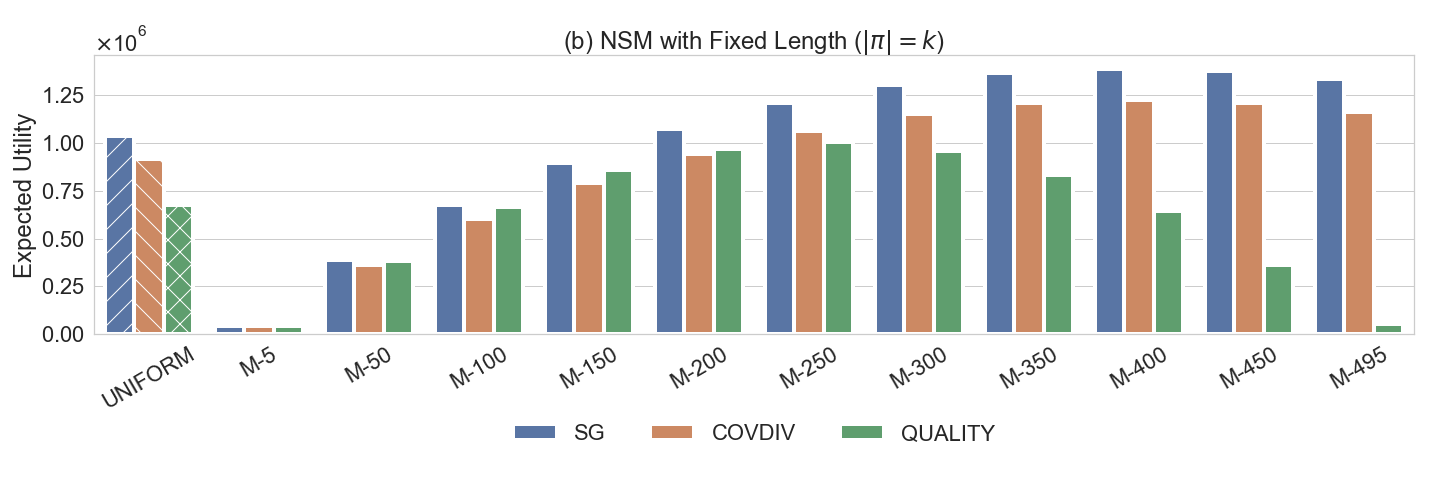}
\vspace*{-0.3in}
\caption{SG achieves superior utility among all three algorithms under various user type distributions.}
\label{fig:sg}
\end{center}
\end{figure*}

\textbf{Algorithms and parameters.} We compare our proposed sampling-based greedy algorithms (labeled as \textsf{SG}) against two baselines, namely, \textsf{COVDIV} and \textsf{QUALITY}, under both flexible length and fixed length settings. \textsf{COVDIV} iteratively selects an item with the largest marginal utility in $g_j(\pi_{[j]})$, i.e., the marginal relevance inspired objective, until no more items with positive marginal utility can be found. \textsf{COVDIV} returns a sequence of items ranked in the same order as they are selected. \textsf{QUALITY} is a simple ranking method that orders individual items in non-increasing quality. Here quality can be measured by average ratings or scores predicted by the recommender systems. Due to space limitation and the results being similar for these metrics, we report the results with average rating metric for \textsf{QUALITY}. For user type distribution $\mathcal{D}$, we use the normal mass function that approximates the normal distribution $\mathcal{N}(\mu,\sigma^2)$ to set the values for $\lambda_j, j\in[k]$. We explore the impact of user type distribution on the performance of the algorithms by varying the value of $\mu$ and $\sigma$.

Each movie, denoted as $i$, is linked to a tag vector $t^i\in[0, 1]^{1128}$. Within this vector, each component represents the relevance score for an individual tag. We employ a widely accepted model to quantify the similarity $w_{ij}$ between two videos, $i$ and $j$, defined as $w_{ij}=\sqrt{\sum_{l=1}^{1128}(\min\{t_l^i,t_l^j\})^2}$ \citep{amanatidis2020fast}. This metric calculates the L2 norm of the element-wise minimum between $t^i$ and $t^j$.  We set $\eta=35$ and adjust the parameters $\alpha$ and $\beta$ to ensure that the two components in $f_j(\cdot)$ are roughly equal in magnitude. In each experimental set, we conduct $100$ rounds, and the average results are presented as follows.

\textbf{Experimental results.} We measure the performance of the algorithms in terms of their expected utility with respect to various user type distributions. As shown in Figure \ref{fig:sg}, we test under a uniform user type distribution where $\lambda_j=1/k, j\in[k]$, labeled as UNIFORM on the $x$-axis. We also report the results under the approximated normal distribution with varying mean, $\mu$, labeled as M-$\mu$ on the $x$-axis. In order to distinguish these two types of distributions, we add hatches on the bars for the uniform distribution. In our experiments, we set $k=500$. Figure \ref{fig:sg}(a) and (b) show the results for NSM with flexible length, i.e., $|\pi|\leq k$, and that for NSM with fixed length, i.e., $|\pi|=k$, respectively.

It shows in Figure \ref{fig:sg}(a) that \textsf{SG} outperforms the benchmarks under all tested user type distributions. While the benchmarks yield an expected utility of $7.29\times 10^5$ and $6.75\times 10^5$ respectively under the uniform distribution, \textsf{SG} yields an expected utility over $1.04\times 10^6$, a $43\%$ increase. Under the approximated normal distribution, the increase of $\mu$ indicates a higher number of videos viewed by average users, leading to an increase in the expected utility for the algorithms. \textsf{QUALITY} always returns a sequence of size $k$ as adding more videos always increases the sum of ratings. However, our objective function is non-monotone, similar videos added by \textsf{QUALITY} result in a lower expected utility as $\mu$ further increases. \textsf{SG} and \textsf{COVDIV} only add items with positive marginal utility. In our experiments, \textsf{SG} returns a sequence of around $400$ videos, and \textsf{COVDIV} returns around $200$ videos. As $\mu$ further increases, their overall expected utility converge since most users will view all the videos listed. We observe that \textsf{SG} outperforms both benchmarks under all test settings. While \textsf{QUALITY} solely considers the ratings of the items, \textsf{COVDIV} only considers their capability of representing the whole collection. \textsf{SG} shows a superior balance between the two. This result validates the superiority of our proposed algorithm over the benchmarks.

Figure \ref{fig:sg}(b) illustrates the results for NSM with fixed length, which means all three algorithms return a sequence of $500$ videos. The results for \textsf{QUALITY} remain the same. We observe that under the uniform distribution, \textsf{SG} yields a lower expected utility while \textsf{COVDIV} yields a higher one, compared with the case of flexible length. We also observe that under approximated normal distribution, the expected utility of \textsf{SG} starts to decrease as $\mu$ goes over $400$, due to the negative contribution from the lastly added videos. The expected utility of \textsf{COVDIV} peaks at $\mu=250$, and then declines. The reason is that for $200<\mu<250$, the increase in the total rating is enough to compensate for the negative contribution from $g_j(\cdot)$, which does not hold any more as $\mu$ further increases. Notably, \textsf{SG} outperforms the others, underscoring our method's advantage.

\bibliographystyle{ijocv081}
\bibliography{reference}

\begin{thebibliography}{23}
\expandafter\ifx\csname natexlab\endcsname\relax\def\natexlab#1{#1}\fi
\expandafter\ifx\csname url\endcsname\relax
  \def\url#1{{\tt #1}}\fi
\expandafter\ifx\csname urlprefix\endcsname\relax\def\urlprefix{URL }\fi
\expandafter\ifx\csname urlstyle\endcsname\relax
  \expandafter\ifx\csname doi\endcsname\relax
  \def\doi#1{doi:\discretionary{}{}{}#1}\fi \else
  \expandafter\ifx\csname doi\endcsname\relax
  \def\doi{doi:\discretionary{}{}{}\begingroup \urlstyle{rm}\Url}\fi \fi

\bibitem[{Alaei et~al.(2010)Alaei, Makhdoumi, and
  Malekian}]{alaei2010maximizing}
Alaei, Saeed, Ali Makhdoumi, Azarakhsh Malekian. 2010.
\newblock Maximizing sequence-submodular functions and its application to
  online advertising.
\newblock {\it arXiv preprint arXiv:1009.4153\/} .

\bibitem[{Amanatidis et~al.(2020)Amanatidis, Fusco, Lazos, Leonardi, and
  Reiffenh{\"a}user}]{amanatidis2020fast}
Amanatidis, Georgios, Federico Fusco, Philip Lazos, Stefano Leonardi, Rebecca
  Reiffenh{\"a}user. 2020.
\newblock Fast adaptive non-monotone submodular maximization subject to a
  knapsack constraint.
\newblock {\it Advances in neural information processing systems\/}.

\bibitem[{Asadpour et~al.(2022)Asadpour, Niazadeh, Saberi, and
  Shameli}]{asadpour2022sequential}
Asadpour, Arash, Rad Niazadeh, Amin Saberi, Ali Shameli. 2022.
\newblock Sequential submodular maximization and applications to ranking an
  assortment of products.
\newblock {\it Operations Research\/} .

\bibitem[{Azar and Gamzu(2011)}]{azar2011ranking}
Azar, Yossi, Iftah Gamzu. 2011.
\newblock Ranking with submodular valuations.
\newblock {\it Proceedings of the twenty-second annual ACM-SIAM symposium on
  Discrete Algorithms\/}. SIAM, 1070--1079.

\bibitem[{Buchbinder and Feldman(2019)}]{buchbinder2019constrained}
Buchbinder, Niv, Moran Feldman. 2019.
\newblock Constrained submodular maximization via a nonsymmetric technique.
\newblock {\it Mathematics of Operations Research\/} {\bf 44} 988--1005.

\bibitem[{Buchbinder et~al.(2014)Buchbinder, Feldman, Naor, and
  Schwartz}]{buchbinder2014submodular}
Buchbinder, Niv, Moran Feldman, Joseph Naor, Roy Schwartz. 2014.
\newblock Submodular maximization with cardinality constraints.
\newblock {\it Proceedings of the twenty-fifth annual ACM-SIAM symposium on
  Discrete algorithms\/}. SIAM, 1433--1452.

\bibitem[{Carbonell and Goldstein(1998)}]{carbonell1998use}
Carbonell, Jaime, Jade Goldstein. 1998.
\newblock The use of mmr, diversity-based reranking for reordering documents
  and producing summaries.
\newblock {\it Proceedings of the 21st annual international ACM SIGIR
  conference on Research and development in information retrieval\/}. 335--336.

\bibitem[{Das and Kempe(2008)}]{das2008algorithms}
Das, Abhimanyu, David Kempe. 2008.
\newblock Algorithms for subset selection in linear regression.
\newblock {\it Proceedings of the fortieth annual ACM symposium on Theory of
  computing\/}. 45--54.

\bibitem[{Das and Kempe(2011)}]{das2011submodular}
Das, Abhimanyu, David Kempe. 2011.
\newblock Submodular meets spectral: greedy algorithms for subset selection,
  sparse approximation and dictionary selection.
\newblock {\it Proceedings of the 28th International Conference on
  International Conference on Machine Learning\/}. 1057--1064.

\bibitem[{Gharan and Vondr{\'a}k(2011)}]{gharan2011submodular}
Gharan, Shayan~Oveis, Jan Vondr{\'a}k. 2011.
\newblock Submodular maximization by simulated annealing.
\newblock {\it Proceedings of the twenty-second annual ACM-SIAM symposium on
  Discrete Algorithms\/}. SIAM, 1098--1116.

\bibitem[{Golovin and Krause(2011)}]{golovin2011adaptive}
Golovin, Daniel, Andreas Krause. 2011.
\newblock Adaptive submodularity: Theory and applications in active learning
  and stochastic optimization.
\newblock {\it Journal of Artificial Intelligence Research\/} {\bf 42}
  427--486.

\bibitem[{Gotovos et~al.(2015)Gotovos, Karbasi, and Krause}]{gotovos2015non}
Gotovos, Alkis, Amin Karbasi, Andreas Krause. 2015.
\newblock Non-monotone adaptive submodular maximization.
\newblock {\it Twenty-Fourth International Joint Conference on Artificial
  Intelligence\/}.

\bibitem[{Harper and Konstan(2015)}]{harper2015movielens}
Harper, F~Maxwell, Joseph~A Konstan. 2015.
\newblock The movielens datasets: History and context.
\newblock {\it Acm transactions on interactive intelligent systems (tiis)\/}
  {\bf 5} 1--19.

\bibitem[{Lin and Bilmes(2010)}]{lin2010multi}
Lin, Hui, Jeff Bilmes. 2010.
\newblock Multi-document summarization via budgeted maximization of submodular
  functions.
\newblock {\it Human Language Technologies: The 2010 Annual Conference of the
  North American Chapter of the Association for Computational Linguistics\/}.
  912--920.

\bibitem[{Lin and Bilmes(2011)}]{lin2011class}
Lin, Hui, Jeff Bilmes. 2011.
\newblock A class of submodular functions for document summarization.
\newblock {\it Proceedings of the 49th Annual Meeting of the Association for
  Computational Linguistics: Human Language Technologies\/}. 510--520.

\bibitem[{Mirzasoleiman et~al.(2016)Mirzasoleiman, Badanidiyuru, and
  Karbasi}]{mirzasoleiman2016fast}
Mirzasoleiman, Baharan, Ashwinkumar Badanidiyuru, Amin Karbasi. 2016.
\newblock Fast constrained submodular maximization: Personalized data
  summarization.
\newblock {\it ICML\/}. 1358--1367.

\bibitem[{Tang(2021)}]{tang2021beyond}
Tang, Shaojie. 2021.
\newblock Beyond pointwise submodularity: Non-monotone adaptive submodular
  maximization in linear time.
\newblock {\it Theoretical Computer Science\/} {\bf 850} 249--261.

\bibitem[{Tang and Yuan(2020)}]{tang2020influence}
Tang, Shaojie, Jing Yuan. 2020.
\newblock Influence maximization with partial feedback.
\newblock {\it Operations Research Letters\/} {\bf 48} 24--28.

\bibitem[{Tang and Yuan(2021{\natexlab{a}})}]{tang2021adaptive}
Tang, Shaojie, Jing Yuan. 2021{\natexlab{a}}.
\newblock Adaptive regularized submodular maximization.
\newblock {\it 32nd International Symposium on Algorithms and Computation
  (ISAAC 2021)\/}. Schloss Dagstuhl-Leibniz-Zentrum f{\"u}r Informatik.

\bibitem[{Tang and Yuan(2021{\natexlab{b}})}]{tang2021cascade}
Tang, Shaojie, Jing Yuan. 2021{\natexlab{b}}.
\newblock Cascade submodular maximization: Question selection and sequencing in
  online personality quiz.
\newblock {\it Production and Operations Management\/} {\bf 30} 2143--2161.

\bibitem[{Tang and Yuan(2022)}]{tang2021optimal}
Tang, Shaojie, Jing Yuan. 2022.
\newblock Optimal sampling gaps for adaptive submodular maximization.
\newblock {\it AAAI\/}.

\bibitem[{Tschiatschek et~al.(2017)Tschiatschek, Singla, and
  Krause}]{tschiatschek2017selecting}
Tschiatschek, Sebastian, Adish Singla, Andreas Krause. 2017.
\newblock Selecting sequences of items via submodular maximization.
\newblock {\it Thirty-First AAAI Conference on Artificial Intelligence\/}.

\bibitem[{Zhang et~al.(2022)Zhang, Tatti, and Gionis}]{zhang2022ranking}
Zhang, Guangyi, Nikolaj Tatti, Aristides Gionis. 2022.
\newblock Ranking with submodular functions on a budget.
\newblock {\it Data mining and knowledge discovery\/} {\bf 36} 1197--1218.

\end{thebibliography}

\section{Appendix: Missing Proofs}

\begin{lemma}
\label{lem:aaa}
\begin{eqnarray}
p\cdot  \mathbb{E} [F(\pi \uplus \pi^*_{\textsf{cons.}})- F(\pi)] \leq \mathbb{E} [F(\pi)].
\end{eqnarray}
\end{lemma}
\emph{Proof:} Let's consider an item $i \in \pi^*$ in the optimal solution and define the events $\mathcal{E}_i$ that capture the sequence of actions taken by our algorithm until item $i$ is considered. Let $q$ denote the position of $i$ in $\pi^*$, that is, $i=\pi^*_{q}$, and let $\mathcal{F}_{i}$ represent the event that $i$ is in $O_{\textsf{cons.}}$, that is, $i\in O_{\textsf{cons.}}$.

Furthermore, let $\Delta_i$ denote the marginal contribution of item $i$ to our solution $\pi$. Given $\mathcal{E}_i$, we define $\pi^{i}$ as the partial solution of $\pi$ right before item $i$ is being considered. Observe that
\begin{eqnarray}
 &&\mathbb{E} [\Delta_i \mid \mathcal{E}_i] =   \mathbb{E} [ \sum_{j\in \{|\pi^i|+1, \cdots, k\}}p\cdot \lambda_j f_j (i \mid \pi^i) \mid \mathcal{E}_i] \\
  &&\geq   \mathbb{E} [ \sum_{j\in \{q, \cdots, k\}}p\cdot \lambda_j  f_j (i \mid \pi^i) \mid \mathcal{E}_i \& \mathcal{F}_{i}]\\
    &&=   p\cdot  \mathbb{E} [ \sum_{j\in \{q, \cdots, k\}}\lambda_j  f_j (i \mid \pi^i) \mid \mathcal{E}_i \& \mathcal{F}_{i}]\\
 &&\geq p\cdot \mathbb{E} [ \sum_{j\in\{q, \cdots, k\}} \lambda_j  f_j(i \mid \pi_{[j]}) \mid \mathcal{E}_i \& \mathcal{F}_{i}]\\
 &&\geq  p\cdot \mathbb{E} [ \sum_{j\in\{q, \cdots, k\}} \lambda_j  f_j(i \mid \pi_{[j]} \cup(\pi^*_{[q-1]}\cap O_{\textsf{cons.}})) \mid \mathcal{E}_i \& \mathcal{F}_{i}]
\end{eqnarray}
where the first inequality is by the observations that  $|\pi^i|+1\leq q$ (this is because if $i\in O_{\textsf{cons.}}$, then $i$ must be considered before its position in $\pi^*$ (e.g., position $q$)); the second inequality is because $f_j$ is submodular, $\pi^i\subseteq \pi_{[j]}$ and $i\notin \pi_{[j]}$ for all $j\geq q$ (because $i\in O_{\textsf{cons.}}$ which implies that $i$ has not been added to $\pi$, we have $i\notin \pi_{[j]}$); the last inequality is because $f_j$ is submodular, $ \pi_{[j]} \subseteq \pi_{[j]} \cup(\pi^*_{[q-1]}\cap O_{\textsf{cons.}})$ and $i\notin \pi_{[j]} \cup(\pi^*_{[q-1]}\cap O_{\textsf{cons.}})$ (this is because  $i\notin \pi_{[j]}$ and $i\notin \pi^*_{[q-1]}$).

Summing up the aforementioned inequality across all $i\in \pi^*$, and employing the law of total probability with respect to $\mathcal{E}_i$ and $\mathcal{F}_{i}$, we can conclude that the total contribution made by items in $\pi^*$ to our solution is at least $p\cdot  \mathbb{E} [F(\pi \uplus \pi^*_{\textsf{cons.}})- F(\pi)]$, that is, $\sum_{i\in\pi^*}\mathbb{E} [\Delta_i]  \geq p\cdot  \mathbb{E} [F(\pi \uplus \pi^*_{\textsf{cons.}})- F(\pi)]$. Hence, the overall utility of our solution  $\mathbb{E} [F(\pi)]$, which can be represented as $\sum_{i\in V} \mathbb{E} [\Delta_i]$,  is at least $p\cdot  \mathbb{E} [F(\pi \uplus \pi^*_{\textsf{cons.}})- F(\pi)]$, observing that $\sum_{i\in V} \mathbb{E} [\Delta_i]\geq \sum_{i\in\pi^*}\mathbb{E} [\Delta_i]  \geq p\cdot  \mathbb{E} [F(\pi \uplus \pi^*_{\textsf{cons.}})- F(\pi)]$. $\Box$
\begin{lemma}
\label{lem:bbb}
\begin{eqnarray}
\mathbb{E} [F(\pi \uplus \pi^*_{\textsf{not cons.}})- F(\pi)]\leq \mathbb{E} [F(\pi)].
\end{eqnarray}
\end{lemma}
\emph{Proof:} Consider a fixed realization of the random coin tosses, the corresponding $\pi$ and $O_{\textsf{not cons.}}$, the following chain, together with the law of total probability, proves this lemma:
\begin{eqnarray}
&&F(\pi) = \sum_{t\in[k]} \sum_{j\in\{t, \cdots, k\}} \lambda_j\cdot f_j(\pi_t \mid \pi_{[t-1]})\\
&&\geq \sum_{t\in[k]} \sum_{j\in\{t, \cdots, k\}} \lambda_j\cdot \frac{1}{2}\cdot f_j(\pi_t \cup (\pi^*_t\cap O_{\textsf{not cons.}}) \mid \pi_{[t-1]}\cup(\pi^*_{[t-1]}\cap O_{\textsf{not cons.}}))\label{eq:line}\\
&&= \frac{1}{2}\cdot \sum_{t\in[k]} \sum_{j\in\{t, \cdots, k\}} \lambda_j\cdot  f_j(\pi_t \cup (\pi^*_t\cap O_{\textsf{not cons.}}) \mid \pi_{[t-1]}\cup(\pi^*_{[t-1]}\cap O_{\textsf{not cons.}}))\\
&& =\frac{1}{2}\cdot  \sum_{j\in[k]}\lambda_j\cdot f_j(\pi_{[j]}\cup(\pi^*_{[j]}\cap O_{\textsf{not cons.}}))=\frac{1}{2}\cdot F(\pi \uplus \pi^*_{\textsf{not cons.}}).
\end{eqnarray}

To prove inequality (\ref{eq:line}), it suffices to show that for all $t\in [k]$,
\begin{eqnarray}
&& \sum_{j\in\{t, \cdots, k\}} \lambda_j\cdot f_j(\pi_t \mid \pi_{[t-1]})~\nonumber\\
&&\geq \sum_{j\in\{t, \cdots, k\}} \lambda_j\cdot \frac{1}{2}\cdot f_j(\pi_t \cup (\pi^*_t\cap O_{\textsf{not cons.}}) \mid \pi_{[t-1]}\cup(\pi^*_{[t-1]}\cap O_{\textsf{not cons.}})).\label{eq:7}
\end{eqnarray}

The rest of the proof is devoted to proving this inequality. First, we show that for all $t\in [k]$,
\begin{eqnarray}
&& \sum_{j\in\{t, \cdots, k\}} \lambda_j\cdot f_j(\pi_t \mid \pi_{[t-1]})\\
&&\geq \sum_{j\in\{t, \cdots, k\}} \lambda_j\cdot f_j(\pi_t  \mid \pi_{[t-1]}\cup(\pi^*_{[t-1]}\cap O_{\textsf{not cons.}})).\label{eq:6}
\end{eqnarray}
To prove this inequality, we first consider the trivial case when $\pi_t=\emptyset$, that is, the actual size of $\pi$ is less than $t$. In this case, we have $\sum_{j\in\{t, \cdots, k\}} \lambda_j\cdot f_j(\pi_t \mid \pi_{[t-1]})= \sum_{j\in\{t, \cdots, k\}} \lambda_j\cdot f_j(\pi_t  \mid \pi_{[t-1]}\cup(\pi^*_{[t-1]}\cap O_{\textsf{not cons.}}))=0$. We next focus on the case when $\pi_t\neq\emptyset$. Let us consider two cases: (1) if $\pi_t\in \pi_{[t-1]}\cup(\pi^*_{[t-1]}\cap O_{\textsf{not cons.}})$, then  $\sum_{j\in\{t, \cdots, k\}} \lambda_j\cdot f_j(\pi_t  \mid \pi_{[t-1]}\cup(\pi^*_{[t-1]}\cap O_{\textsf{not cons.}}))=0$. This, together with the observation that $\sum_{j\in\{t, \cdots, k\}} \lambda_j\cdot f_j(\pi_t \mid \pi_{[t-1]})>0$ (this is by the design of our algorithm, i.e., we only add items that bring positive marginal utility to our solution), implies  $\sum_{j\in\{t, \cdots, k\}} \lambda_j\cdot f_j(\pi_t \mid \pi_{[t-1]})
\geq \sum_{j\in\{t, \cdots, k\}} \lambda_j\cdot f_j(\pi_t  \mid \pi_{[t-1]}\cup(\pi^*_{[t-1]}\cap O_{\textsf{not cons.}}))$. (2) if $\pi_t\notin \pi_{[t-1]}\cup(\pi^*_{[t-1]}\cap O_{\textsf{not cons.}})$, then by the assumption that $f_i$ is submodular and the observation that $\pi_{[t-1]}\subseteq \pi_{[t-1]}\cup(\pi^*_{[t-1]}\cap O_{\textsf{not cons.}})$, we have  $\sum_{j\in\{t, \cdots, k\}} \lambda_j\cdot f_j(\pi_t \mid \pi_{[t-1]})
\geq \sum_{j\in\{t, \cdots, k\}} \lambda_j\cdot f_j(\pi_t  \mid \pi_{[t-1]}\cup(\pi^*_{[t-1]}\cap O_{\textsf{not cons.}}))$.

Second, we show that
\begin{eqnarray}
&& \sum_{j\in\{t, \cdots, k\}} \lambda_j\cdot f_j(\pi_t \mid \pi_{[t-1]})\\
&&\geq \sum_{j\in\{t, \cdots, k\}} \lambda_j\cdot f_j(\pi^*_t\cap O_{\textsf{not cons.}} \mid \pi_{[t-1]}\cup(\pi^*_{[t-1]}\cap O_{\textsf{not cons.}})).\label{eq:8ooo}
\end{eqnarray}
To avoid trivial cases, let us assume $\pi^*_t\neq \emptyset$, i.e., $|\pi^*|\geq t$. To prove this inequality, we first consider the case when $\pi^*_t\cap O_{\textsf{not cons.}} =\emptyset$, that is, $\pi^*_t$ does not belong to $O_{\textsf{not cons.}}$. In this case, the above inequality holds because $\sum_{j\in\{t, \cdots, k\}} \lambda_j\cdot f_j(\pi_t \mid \pi_{[t-1]})\geq 0$ (this holds even if $\pi_t=\emptyset$) and $\sum_{j\in\{t, \cdots, k\}} \lambda_j\cdot f_j(\pi^*_t\cap O_{\textsf{not cons.}} \mid \pi_{[t-1]}\cup(\pi^*_{[t-1]}\cap O_{\textsf{not cons.}}))=0$. We next prove the case when  $\pi^*_t\cap O_{\textsf{not cons.}} \neq\emptyset$. That is,  $\pi^*_t\cap O_{\textsf{not cons.}} = \pi^*_t$. Observe that
\begin{eqnarray}
&& \sum_{j\in\{t, \cdots, k\}} \lambda_j\cdot f_j(\pi_t \mid \pi_{[t-1]})\geq \sum_{j\in\{t, \cdots, k\}} \lambda_j\cdot f_j(\pi^*_t \mid \pi_{[t-1]}) \label{eq:4}
\\
&&\geq \sum_{j\in\{t, \cdots, k\}} \lambda_j\cdot f_j(\pi^*_t \mid \pi_{[t-1]}\cup(\pi^*_{[t-1]}\cap O_{\textsf{not cons.}})). \label{eq:5}
\end{eqnarray}
 The first inequality is by the design of our algorithm, that is, we select an item that maximizes the marginal utility on top of $\pi_{[t-1]}$ as the $t$-th item $\pi_t$, hence, every item that has not been considered so far, including $\pi^*_t$, must have marginal utility that is no larger than $\pi_t$. Note that this inequality holds even if $\pi_t=\emptyset$, this is because in this case, by the design of our algorithm, the reason why we did not consider $\pi^*_t$ is because adding $\pi^*_t$ to the current solution  incurs negative (including zero) marginal utility, i.e., $\sum_{j\in\{t, \cdots, k\}} \lambda_j\cdot f_j(\pi^*_t \mid \pi_{[t-1]})\leq 0$, hence,  $\sum_{j\in\{t, \cdots, k\}} \lambda_j\cdot f_j(\pi_t \mid \pi_{[t-1]})=0 \geq \sum_{j\in\{t, \cdots, k\}} \lambda_j\cdot f_j(\pi^*_t \mid \pi_{[t-1]})$. Inequality (\ref{eq:5}) is by the assumption that $f_j$ is submodular, $\pi_{[t-1]}\subseteq  \pi_{[t-1]}\cup(\pi^*_{[t-1]}\cap O_{\textsf{not cons.}})$ and $\pi^*_t\notin  \pi_{[t-1]}\cup(\pi^*_{[t-1]}\cap O_{\textsf{not cons.}})$. Here, $\pi^*_t\notin  \pi_{[t-1]}\cup(\pi^*_{[t-1]}\cap O_{\textsf{not cons.}})$ is by the following two observations: (1) $\pi^*_t\notin  \pi_{[t-1]}$, which is because $\pi^*_t \in O_{\textsf{not cons.}}$, and (2) $\pi^*_t\notin  \pi^*_{[t-1]}\cap O_{\textsf{not cons.}}$, which is because  $\pi^*_t\notin  \pi^*_{[t-1]}$.

Inequalities (\ref{eq:6}) and (\ref{eq:8ooo}) imply that
\begin{eqnarray}
&&2\times  \sum_{j\in\{t, \cdots, k\}} \lambda_j\cdot f_j(\pi_t \mid \pi_{[t-1]})\\
&&\geq \sum_{j\in\{t, \cdots, k\}} \lambda_j\cdot f_j(\pi_t  \mid \pi_{[t-1]}\cup(\pi^*_{[t-1]}\cap O_{\textsf{not cons.}}))\\
 &&\quad\quad\quad+  \sum_{j\in\{t, \cdots, k\}} \lambda_j\cdot f_j(\pi^*_t\cap O_{\textsf{not cons.}} \mid \pi_{[t-1]}\cup(\pi^*_{[t-1]}\cap O_{\textsf{not cons.}}))\\
&&\geq \sum_{j\in\{t, \cdots, k\}} \lambda_j\cdot f_j(\pi_t \cup (\pi^*_t\cap O_{\textsf{not cons.}}) \mid \pi_{[t-1]}\cup(\pi^*_{[t-1]}\cap O_{\textsf{not cons.}}))
\end{eqnarray}
where the second inequality is by the assumption that $f_j$ is submodular. Hence, inequality (\ref{eq:7}) is proved. $\Box$




\end{document}